%% file: impose.tex
\documentclass[journal]{styles/IEEEtran}

\usepackage{graphicx}
\usepackage{amsmath}
\usepackage{amssymb}
\usepackage{multirow}
\usepackage{booktabs}
\usepackage{float}
\usepackage{capt-of}
\usepackage{threeparttable}
\usepackage{makecell}
\usepackage[table]{xcolor}
\input{macros/revision_macros.tex}
\usepackage[caption=false,font=footnotesize]{subfig}
\usepackage{url}
\usepackage{array}
\usepackage{etoolbox}
\usepackage{ragged2e}
\usepackage{placeins}
\usepackage{needspace}
\usepackage{hyperref}
\IfFileExists{orcidlink.sty}{\usepackage{orcidlink}}{\newcommand{\orcidlink}[1]{}}
\usepackage{microtype}

\hypersetup{hidelinks}

\begin{document}

\title{Identity-Consistent Multi-Pose Generation of Contactless Fingerprints}

\author{Zhiyu Pan$^{\orcidlink{0009-0000-6721-4482}}$,
  Zhaoliang Xu$^{\orcidlink{0009-0005-9680-6272}}$,
  Xiongjun Guan$^{\orcidlink{0000-0001-8887-3735}}$,
  Jianjiang Feng$^{\orcidlink{0000-0003-4971-6707}}$, ~\IEEEmembership{Member, IEEE},
  and Jie Zhou$^{\orcidlink{0000-0001-7701-234X}}$, ~\IEEEmembership{Fellow, IEEE}
  \thanks{
    This work was supported in part by the National Natural Science Foundation of China under Grant 62376132 and 62321005. (\emph{Corresponding author: Jianjiang Feng}.)}
  \IEEEcompsocitemizethanks{
  \IEEEcompsocthanksitem
  Zhiyu Pan, Xiongjun Guan, Jianjiang Feng, and Jie Zhou are with the Department of Automation, Tsinghua University, Beijing 100084, China (e-mail: \url{pzy20@tsinghua.org.cn}; \url{gxj21@mails.tsinghua.edu.cn};
  \url{jfeng@tsinghua.edu.cn}; \url{jzhou@tsinghua.edu.cn}).
  \IEEEcompsocthanksitem
  Zhaoliang Xu is with People's Public Security University of China, Beijing, China (e-mail: \url{hiahiabiu@163.com}).}
  \thanks{This paper includes supplementary material at the end of the document, after the references. It contains autoencoder objectives and architectures, angle-stratified identity evaluation of the projected images, training protocols of the controlled experiments, computational cost, Sauvola sensitivity grids, and additional failure examples. Contact \url{jfeng@tsinghua.edu.cn} for further questions about this work.}}

\markboth{IEEE Transactions on Information Forensics and Security}%
{Pan \MakeLowercase{\textit{et al.}}: Identity-Consistent Multi-Pose Generation of Contactless Fingerprints}

\maketitle

\begin{abstract}
Contactless fingerprint recognition has gained increasing attention due to its advantages in hygiene, acquisition flexibility, and device compatibility. However, the absence of physical contact constraints introduces severe nonlinear geometric distortions caused by free finger poses in 3D space, resulting in a substantial cross-modal domain gap between contactless and conventional contact-based fingerprints. Existing solutions largely rely on explicit geometric correction or image enhancement as preprocessing, but such pipelines are fragile under extreme pose variations. In this paper, we approach the problem from a data-centric perspective and propose Identity-Consistent Multi-Pose Generation of Contactless Fingerprints (IMPOSE), a physics-inspired framework that synthesizes identity-preserving, multi-pose contactless fingerprint samples to train robust recognition models. IMPOSE consists of three core stages: (1) rolled fingerprint identity generation via latent diffusion with discrete codebook representations, (2) cross-modal translation from rolled to contactless modality guided by Sauvola-based local adaptive binarization as an identity anchor, and (3) physics-based multi-pose simulation through 3D finger model texture mapping and projection. {Across all 15,136 final 3D-projected views of the 1,712 retained identities, DMD and VeriFinger achieve 98.57\% and 98.64\% Rank-1 identification, with errors concentrated at roll angles beyond $\pm30^\circ$.} Extensive experiments on the UWA and PolyU CL2CB databases demonstrate that fine-tuning the Fixed-Length Dense Descriptor (FDD) with IMPOSE-synthesized data achieves state-of-the-art cross-modal matching performance, reducing EER to 8.74\% on UWA (all poses) and 2.26\% on PolyU CL2CB. {Fine-tuning with IMPOSE data also consistently improves DeepPrint and AFRNet, with hybrid synthetic-and-real training achieving the best overall performance. Under an equal training budget, 3D-projected views outperform 2D thin-plate-spline augmentation in both Rank-1 and EER on both databases.} Code is available at \url{https://github.com/Yu-Yy/IMPOSE}.
\end{abstract}


\IEEEpeerreviewmaketitle

\newpage
\section{Introduction}
\IEEEPARstart{C}{ontactless} fingerprint recognition has emerged as a prominent research direction owing to its superior hygiene, user convenience, and compatibility with ubiquitous camera-equipped mobile devices~\cite{grosz2021c2cl}. Unlike contact-based acquisition that constrains the finger on a flat platen, contactless capture allows free finger placement in 3D space, reducing shared-surface contact and broadening application scenarios. Fingerprints provide discriminative ridge information, but contactless acquisition also requires attention to presentation attacks and acquisition quality~\cite{maltoni2022handbook}. Despite these advantages, the freedom in finger placement introduces severe multi-pose challenges: rotation, pitch, and yaw of the finger in 3D space produce significant nonlinear geometric distortions and occlusions, creating a substantial cross-modal domain gap between contactless images and conventional contact-based gallery fingerprints.

To address these challenges, mainstream research paradigms have largely focused on image enhancement~\cite{grosz2021c2cl}, cross-modal spatial alignment~\cite{dong2023synthesis,dong2025bridging}, and geometric distortion correction~\cite{cui2023monocular} as preprocessing steps. These methods aim to narrow the domain gap by standardizing contactless images to resemble contact-based counterparts, thereby enabling the extraction of unified discriminative features. However, the inherent complexity of nonlinear finger deformations and the fragility of explicit geometric reconstruction often limit their robustness in extreme pose scenarios.

\begin{figure}[!t]
  \centering
  \includegraphics[width=\linewidth]{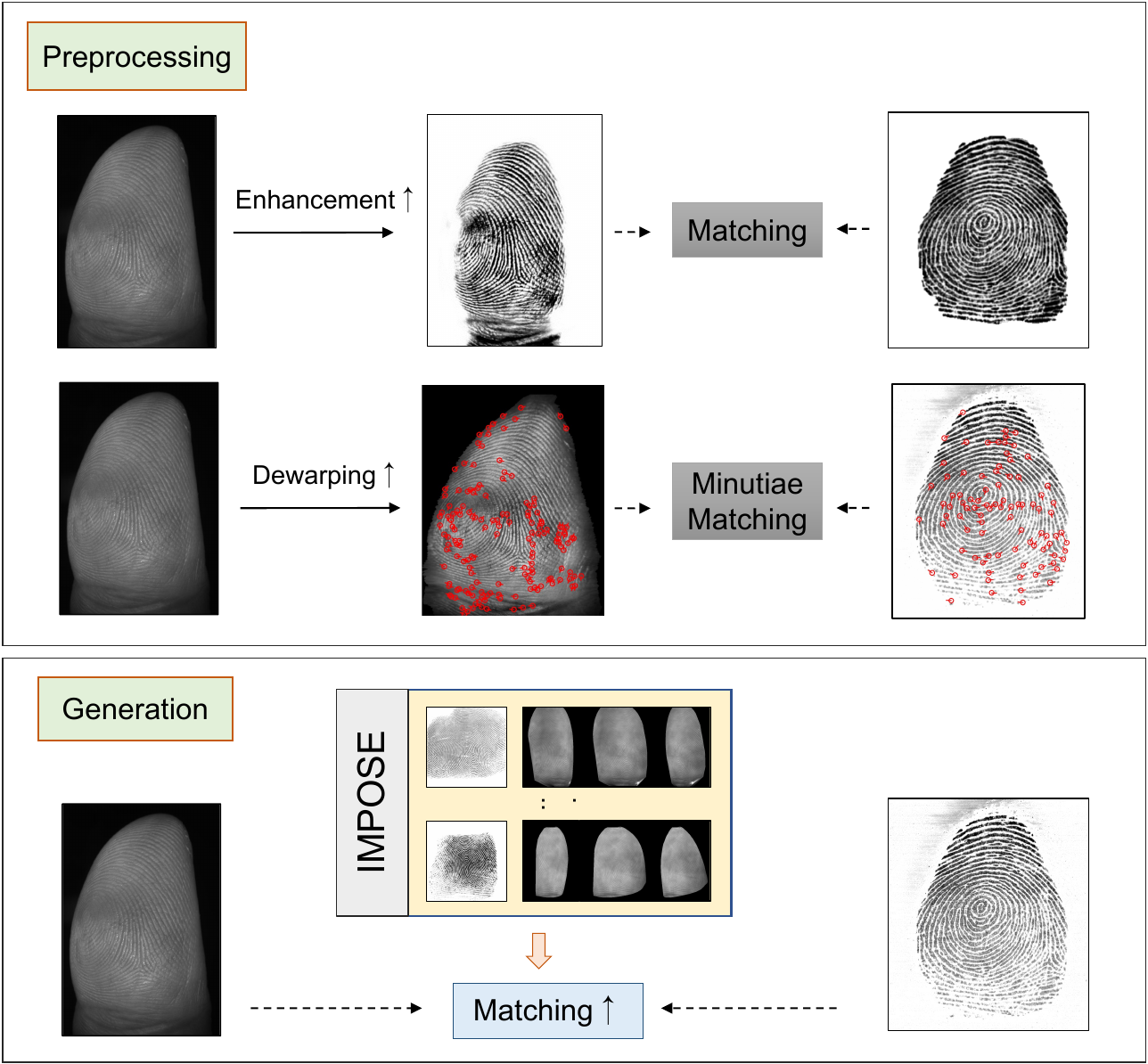}
  \caption{Comparison of strategies for improving contactless fingerprint matching performance. (a) The conventional paradigm relies on explicit geometric correction and enhancement. (b) Our IMPOSE framework addresses the problem from a data-centric perspective, enhancing recognition models through identity-consistent synthetic data.}
  \label{fig:motivation}
\end{figure}

In prior work~\cite{pan2024FDD,pan2025FLARE}, we developed a Fixed-Length Dense Descriptor (FDD) that demonstrated strong robustness in global fingerprint representation. Remarkably, even when trained solely on high-quality rolled fingerprint data, FDD achieved competitive matching performance on multiple heterogeneous fingerprint databases, including PolyU CL2CB~\cite{liu20143d}. This key observation motivates a fundamentally different technical route: rather than designing fragile and complex explicit image reconstruction pipelines, we propose to strengthen the representation network through identity-consistent, multi-pose synthetic samples that cover diverse acquisition conditions. By leveraging physics-simulated contactless multi-modal samples for end-to-end learning, the recognition system can implicitly learn robust features resilient to nonlinear distortions and illumination variations, without relying on cumbersome preprocessing operators (see Fig.~\ref{fig:motivation}).

In fingerprint generation, deep learning has shifted research from hand-crafted statistical modeling~\cite{cappelli2004sfinge} toward data-driven synthesis~\cite{engelsma2022printsgan,grosz2024universal,shoshan2024fpgan}. Recent advances such as GenPrint~\cite{grosz2024universal}, built on denoising diffusion probabilistic models (DDPM)~\cite{ho2020DDPM}, can simulate fingerprints under various sensor styles and text-driven conditions. However, existing methods predominantly focus on conventional contact-based planar or rolled fingerprints, with limited exploration of contactless modalities. Notable exceptions like SynColFinger~\cite{priesnitz2022syncolfinger} rely on traditional SFinGe~\cite{cappelli2004sfinge} texture models, resulting in significant discrepancies in ridge detail and illumination realism. While general-purpose models such as GenPrint can mimic contactless styles, they operate as style transfer on planar fingerprints and cannot physically simulate the nonlinear distortions and multi-pose geometric transformations inherent to unconstrained contactless acquisition.

Motivated by the above observations, we propose Identity-Consistent Multi-Pose Generation of Contactless Fingerprints (IMPOSE), a complete physics-inspired generation pipeline. IMPOSE first synthesizes rolled fingerprints as original identity references, then translates them to the contactless domain through a cross-modal conversion mechanism, and finally simulates diverse acquisition poses and nonlinear distortions via 3D finger point cloud projection. For identity consistency control, we introduce the Sauvola local adaptive thresholding algorithm~\cite{sauvola2000adaptive} as a cross-modal condition. Unlike deep learning-based binarization methods that may produce hallucinated textures, Sauvola is a deterministic digital image processing technique that stably extracts ridge structures in a non-learning manner, with identity preservation evaluated quantitatively.

Using large-scale datasets generated by IMPOSE, we fine-tune several representative fixed-length fingerprint descriptor extraction networks to enhance their matching robustness in contactless scenarios. Extensive quantitative and qualitative experiments validate the richness of generated samples and their ridge-level identity consistency. Verification on the multi-pose contactless database UWA~\cite{zhou2014benchmark} and the cross-modal database PolyU CL2CB~\cite{liu20143d} {shows that fine-tuning with IMPOSE data improves fixed-length representations, with the largest gains at large roll angles.}

The main contributions of this work are:
\begin{enumerate}
    \item We propose IMPOSE, a data-centric framework that approaches contactless fingerprint matching from a generation-and-augmentation perspective. By synthesizing identity-consistent, multi-pose contactless samples for augmentation, IMPOSE enables fixed-length representations to learn robustness against nonlinear distortions, avoiding the fragility of explicit geometric correction.
    \item To preserve identity fidelity during cross-modal generation, we introduce Sauvola local adaptive binarization as a deterministic, non-learned identity anchor that avoids hallucination artifacts. To simulate realistic multi-pose distortion, we construct a physics-based simulation pipeline using 3D finger model texture mapping and projection, using measured finger surfaces to model roll variation.
    \item {Experiments on UWA and PolyU CL2CB show that IMPOSE data consistently improves FDD, DeepPrint, and AFRNet. We further verify identity preservation of the final projected images across roll angles and demonstrate the benefit of 3D projection over 2D thin-plate-spline augmentation under an equal training budget.}
\end{enumerate}

\section{Related Work}

\begin{figure*}[!t]
  \centering
  \includegraphics[width=.95\textwidth]{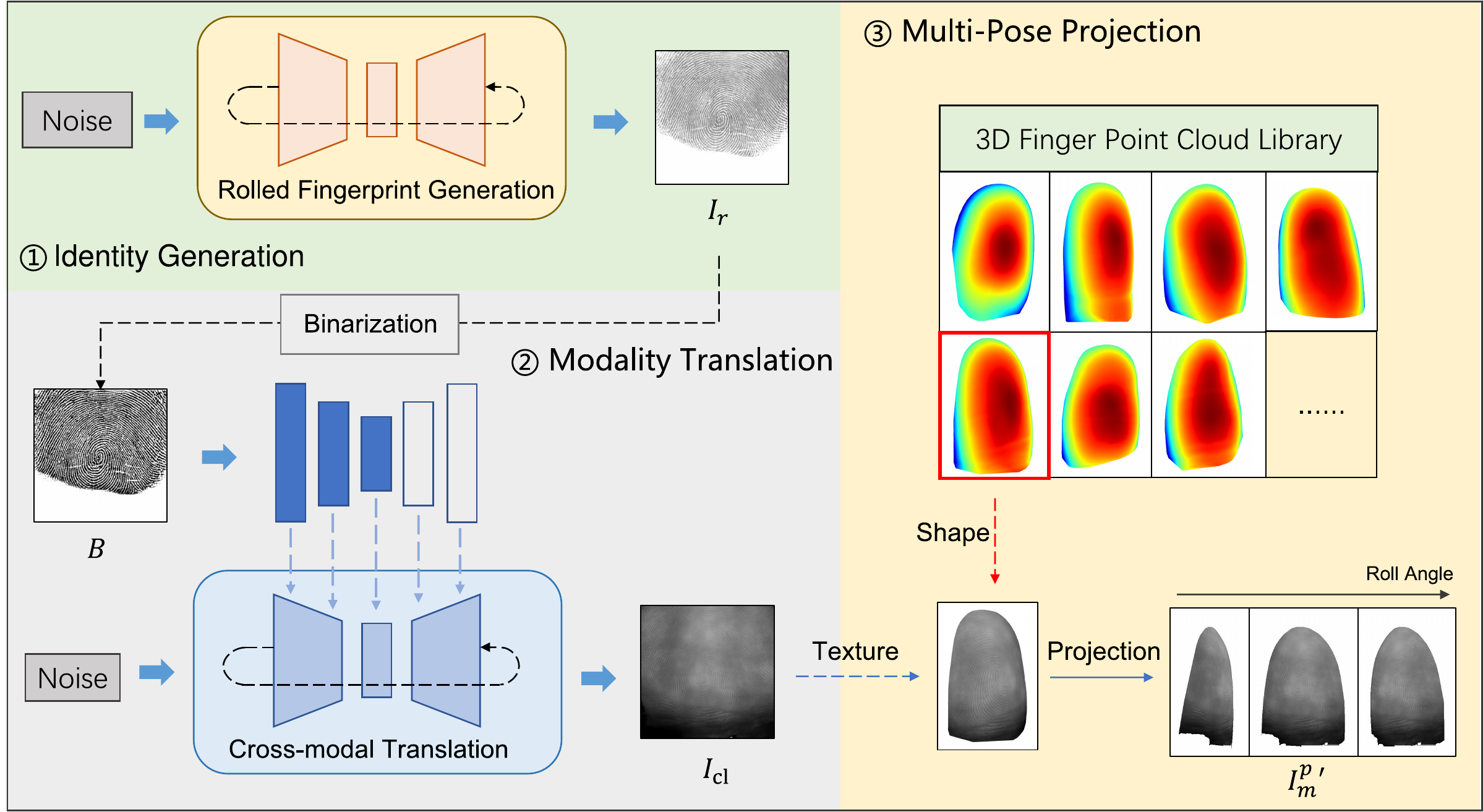}
  \caption{Overview of IMPOSE: identity generation, modality translation and multi-pose projection.}
  \label{fig:pipeline}
\end{figure*}

\subsection{Deep Generative Models for Fingerprint Synthesis}

Fingerprint generation has evolved from physics-based manual modeling to data-driven deep generative approaches. The seminal SFinGe~\cite{cappelli2004sfinge} established a complete physical simulation framework using mathematical ridge flow models with Gabor filters, enabling the generation of multiple impressions from the same virtual finger. Subsequent works introduced statistical modeling of ridge features~\cite{zhao2012fingerprint} and minutiae-guided generation~\cite{johnson2013texture}, but remained constrained by their hand-crafted nature, producing relatively homogeneous patterns lacking complex intra-class variations.

With the rise of deep learning, Generative Adversarial Networks (GANs)~\cite{goodfellow2020GAN} became the dominant paradigm for fingerprint synthesis. Early GAN-based methods~\cite{mistry2020fingerprint,minaee2018finger,riazi2020synfi,bahmani2021high,cao2018fingerprint} significantly improved texture realism but lacked explicit control over fingerprint identity. PrintsGAN~\cite{engelsma2022printsgan} introduced a two-stage generation scheme that first synthesizes a binary ``Master Print'' as an identity reference and then simulates distinct impressions through latent-variable-driven deformation. This decoupling of identity and appearance inspired subsequent work. FPGAN-Control~\cite{shoshan2024fpgan} combined VAE and GAN architectures for fine-grained appearance control through latent space interpolation.

More recently, Denoising Diffusion Probabilistic Models (DDPM)~\cite{ho2020DDPM} have enabled controllable fingerprint generation. GenPrint~\cite{grosz2024universal} exploited DDPM's text-guided control capabilities to generate fingerprints conditioned on human-interpretable attributes including pattern type, sensor type, and quality level, while maintaining identity consistency through a pre-trained ridge extractor.

Despite these advances, existing methods predominantly target contact-based fingerprints. For contactless modalities, GenPrint~\cite{grosz2024universal} provides preliminary contactless-style generation but operates as style transfer on planar fingerprints without modeling 3D pose variations. SynColFinger~\cite{priesnitz2022syncolfinger} overlays contactless color statistics on SFinGe-generated ridges with hand-crafted distortion fields, resulting in low texture fidelity. Dong et al.~\cite{dong2023synthesis} explored multi-view 3D fingerprint simulation by mapping SFinGe textures onto Bezier-fitted 3D finger surfaces, but the generated textures remain overly idealized. Building a framework that achieves both high-fidelity texture generation and physically-realistic multi-pose distortion simulation remains an open challenge.

\subsection{Contactless Fingerprint Matching and Recognition}

Contactless fingerprint matching has been approached from two main angles: local feature modeling and geometric distortion correction. For local feature modeling, Kumar et al.~\cite{kumar2013towards} proposed 3D finger surface codes extending minutiae into 3D space. Tan et al.~\cite{tan2020towards} employed elliptical finger models with singular point information for pose estimation and improved minutiae extraction. Shi et al.~\cite{shi2022towards} explored CNN- and GCN-based architectures for robust 2D minutiae representation. Dong et al.~\cite{dong2025bridging} achieved robust recovery of 3D minutiae from single 2D contactless images, bridging the modality gap between contactless 3D features and contact-based 2D templates.

For geometric correction, Cui et al.~\cite{cui2023monocular} and Jia et al.~\cite{jia2024improving} predicted finger depth from monocular images and performed unwarping before feature matching. Grosz et al.~\cite{grosz2021c2cl} proposed C2CL, which adaptively learns elastic distortion parameters guided by fixed-length representation similarity, achieving strong cross-modal performance under moderate pose variations.

Camera-acquired fingerprints are also studied as \emph{fingerphotos}: VerifNet~\cite{parasnis2024verifnet} fuses wavelet-based deep features with minutiae matching, LFIW~\cite{liu2024lfiw} benchmarks smartphone, contact, and latent impressions, and UFQA~\cite{joshi2024ufqa} estimates fingerphoto utility. Whereas those works focus on acquisition quality or feature fusion, our approach complements explicit geometric correction by training cross-modal descriptors with multi-pose synthetic samples.

\section{Methodology}

The IMPOSE framework, illustrated in Fig.~\ref{fig:pipeline}, consists of three core stages: (1) rolled fingerprint identity generation, which establishes unique fingerprint identities as the foundational reference; (2) cross-modal translation from rolled to contactless modality guided by Sauvola binary enhancement maps as identity anchors; and (3) physics-based multi-pose simulation through 3D finger model texture mapping and projection. The following subsections detail each stage.

\subsection{Preliminaries: Denoising Diffusion Probabilistic Models}

Denoising Diffusion Probabilistic Models (DDPM)~\cite{ho2020DDPM} are a class of Markov chain-based generative models that transform complex data distributions into simple isotropic Gaussian distributions through simulated diffusion processes.

DDPM consists of a forward diffusion process and a reverse denoising process. The forward process defines a Markov chain by gradually adding Gaussian noise $\epsilon$ to the original data $x_0$ over $T$ timesteps. Using the reparameterization trick, the noisy sample $x_t$ at any timestep $t$ can be directly expressed as:
\begin{equation}
  x_t = \sqrt{\bar{\alpha}_t}x_0 + \sqrt{1 - \bar{\alpha}_t}\epsilon, \quad \epsilon \sim \mathcal{N}(\mathbf{0}, \mathbf{I})
  \label{eq:ddpm_add}
\end{equation}
where $\alpha_t = 1 - \beta_t$ is a predefined hyperparameter and $\bar{\alpha}_t := \prod_{s=0}^{t} \alpha_s$ denotes the cumulative noise coefficient. The training objective is to learn a parameterized network $\epsilon_\theta$ that predicts the noise component added to the data.

To improve computational efficiency for high-resolution image generation, we adopt the Latent Diffusion Model (LDM)~\cite{rombach2022LDM} paradigm, which performs the diffusion process in a low-dimensional latent space constructed by a pretrained variational autoencoder. Under this framework, the encoder $\xi$ maps an input image $x_i$ to a latent representation $z_0 = \xi(x_i)$, allowing the denoising network $\epsilon_\theta$ to operate in a computationally efficient latent space. With cross-modal conditioning constraints {(e.g., a spatial condition)}, the training objective is:
\begin{equation}
  {\mathcal{L}_{\mathrm{diff}} = \mathbb{E}_{z_t, t, C, \epsilon \sim \mathcal{N}(\mathbf{0},\mathbf{I})} \left[ \left\| \epsilon - \epsilon_\theta(z_t, t, C) \right\|_1 \right]}
  \label{eq:ddpm_loss}
\end{equation}
{Here $C=\varnothing$ for unconditional rolled generation and $C=z_B$ (the encoded Sauvola map) for translation. Consistent with our implementation, we use the $\ell_1$ noise-prediction loss instead of the squared $\ell_2$ loss of standard DDPM~\cite{ho2020DDPM}.}

During inference, we adopt the DDIM~\cite{song2020denoising} formulation for iterative denoising from random Gaussian noise $z_T$ to the final latent representation $z_0$. DDIM enables significantly fewer inference steps than training without sacrificing synthesis quality. By setting the stochastic variance term $\sigma_t$ to 0, the reverse process becomes a deterministic mapping:
\begin{equation}
  z_{t-1} = \sqrt{\bar{\alpha}_{t-1}} \hat{z}_0 + \sqrt{1 - \bar{\alpha}_{t-1}} \cdot \epsilon_\theta(z_t, t, C),
  \label{eq:ddim}
\end{equation}
where $\hat{z}_0 = (z_t - \sqrt{1 - \bar{\alpha}_t} \epsilon_\theta(z_t, t, C)) / \sqrt{\bar{\alpha}_t}$ is the predicted clean latent at timestep $t$.

\subsection{Rolled Fingerprint Identity Generation}

The first stage of IMPOSE establishes unique fingerprint identities as the foundational reference for the entire pipeline. We employ a hierarchical approach combining discrete latent space modeling with latent diffusion.

For image compression, we introduce a Vector Quantized Variational Autoencoder (VQ-VAE)~\cite{vqvae} to discretize the fingerprint latent space, constructing a structured codebook. By mapping complex local texture features to finite codebook indices, VQ-VAE achieves precise feature quantization and effective redundancy reduction. This discrete representation captures global topological patterns through modeling the joint distribution of code indices, accurately preserving ridge flow continuity and minutiae uniqueness.

The encoder $E_I$ first maps an input rolled fingerprint image $I$ to continuous latent representations $z_l$. Vector quantization against the codebook yields discrete latent representations $z_i$, which are then decoded by $D_I$ to produce the rolled fingerprint image $I_r$. Considering the complexity of ridge structures, we employ only $4\times$ spatial downsampling with 3 latent channels to prevent irreversible loss of critical spatial information. The codebook size is $N_C = 4,096$ with each feature vector of dimension 3. \rev{The VQ-VAE is pretrained with the sum of an $\ell_1$ reconstruction loss, an LPIPS perceptual loss~\cite{LPIPS}, and the vector-quantization loss, all with unit weight; the commitment coefficient inside the quantization loss is $\beta=0.25$ (Supplementary Table~\suppref{tab:supp_vq_loss}).}

\begin{figure}[!t]
  \centering
  \includegraphics[width=\linewidth]{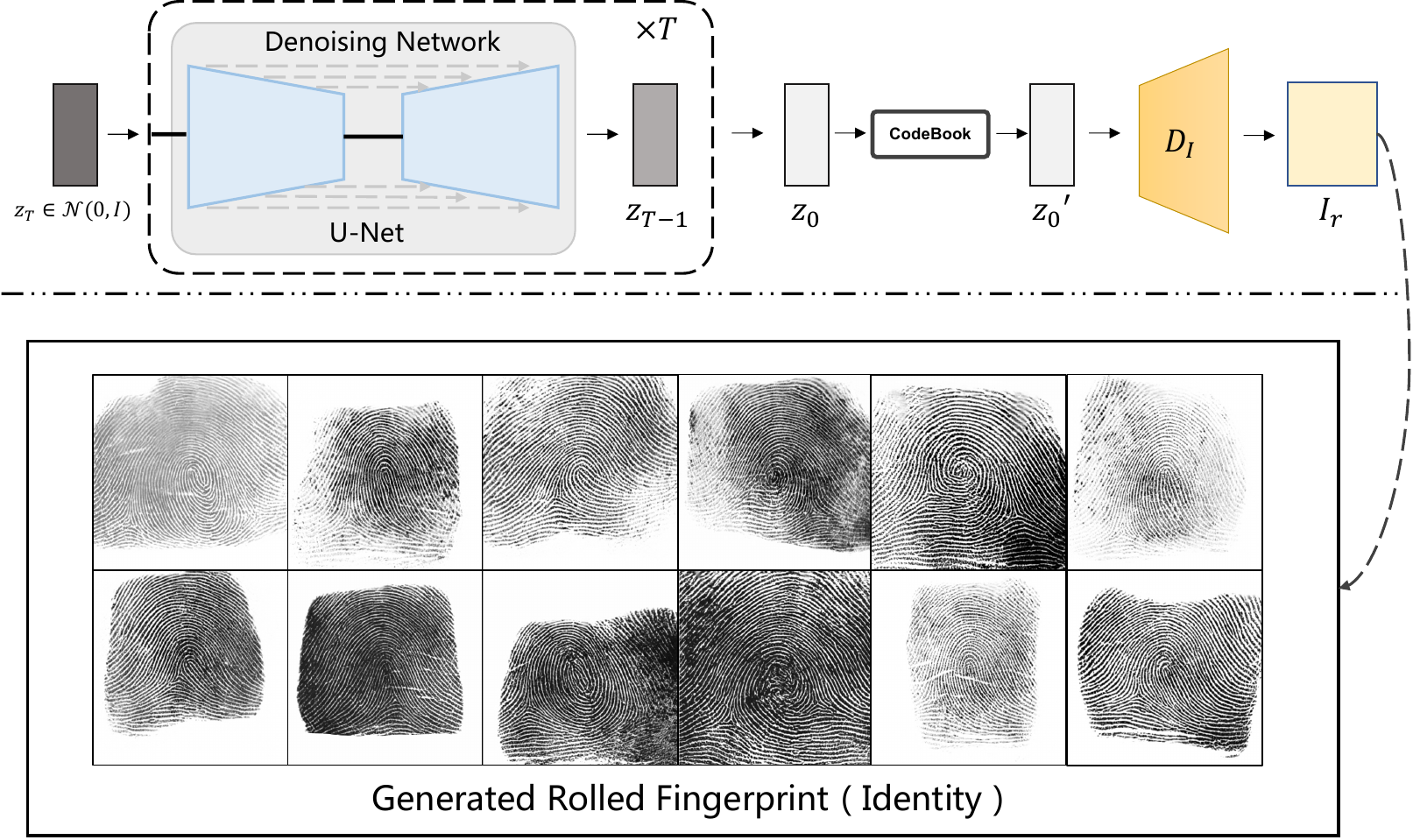}
  \caption{Process of rolled fingerprint synthesis with corresponding generated examples.}
  \label{fig:rolled_gen_pipeline}
\end{figure}

In the diffusion stage, we train a U-Net-based denoising network on the codebook-indexed latent space to synthesize novel fingerprint identities from Gaussian noise. At inference, the system samples random noise $z_T$ in the latent space and iteratively denoises it through $T$ timesteps. To ensure the generated latent vectors respect fingerprint manifold constraints, the derived $\hat{z}_0$ is further refined through codebook-based vector quantization before being decoded to pixel space by the pretrained decoder $D_I$ (see Fig.~\ref{fig:rolled_gen_pipeline}).

The training data consist of 1,093 high-resolution optical rolled fingerprints collected in-house (THU Rolled) and 2,000 rolled fingerprint samples from NIST SD302~\cite{nist302}, totaling 3,093 images. All samples are preprocessed with high-precision pose estimation~\cite{duan2023estimating} for alignment and normalized to $512 \times 512$ pixels at 500 PPI.

\subsection{Identity-Consistent Control via Sauvola Binarization}
\label{sec:sauvola}

The core challenge in rolled-to-contactless modality translation lies in balancing cross-modal style realism with identity invariance. Unlike face recognition focused on global semantics, fingerprint recognition demands extremely low tolerance for ridge and minutiae deviations---subtle alterations in ridge topology can lead to identity shift.

Through empirical investigation, we found that conventional approaches such as fixed-length descriptor supervision~\cite{pan2024FDD}, ridge extractor-based supervision~\cite{nist2020verifinger}, or minutiae-based rigid alignment of training pairs all struggle to achieve effective identity consistency control during generation training, primarily due to inherent differences in acquisition time, contact area, and nonlinear distortion between rolled and contactless fingerprints.

We propose a structural conditioning strategy that uses ridge-level pixel correspondence without an additional identity loss. Drawing inspiration from prior work~\cite{engelsma2022printsgan,grosz2024universal} that uses binary ridge enhancement maps as identity injection signals, we identify two key mechanisms: (1) binary images provide high contrast that clearly characterizes identity features; (2) binary maps extracted directly from the target modality ensure natural ridge-level consistency.

\begin{figure}[!t]
  \centering
  \includegraphics[width=\linewidth]{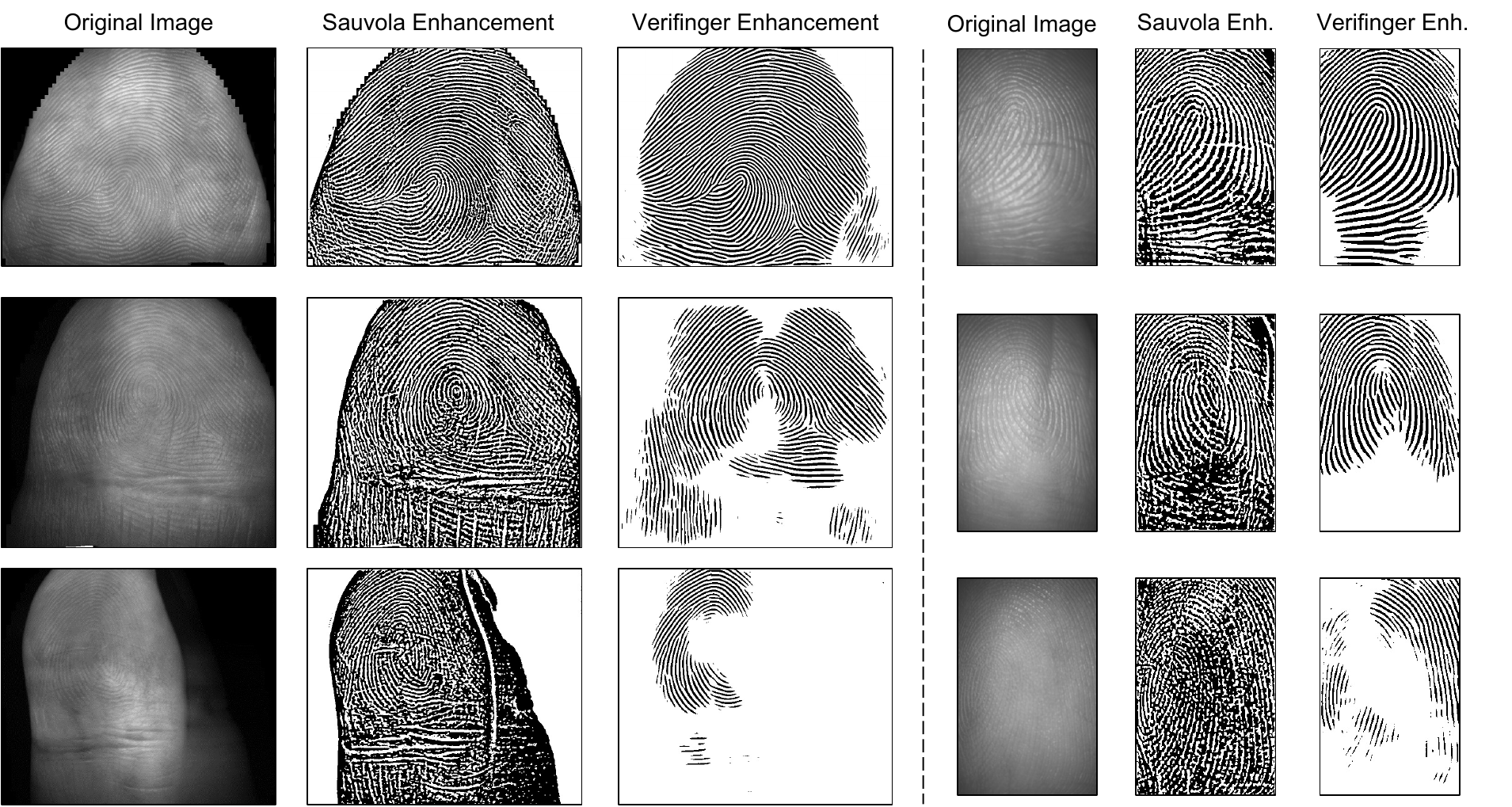}
  \caption{Comparison of Sauvola binarization against VeriFinger v12.0 on contactless fingerprints.}
  \label{fig:enh_samples}
\end{figure}

However, contactless fingerprints typically exhibit low contrast, causing mature ridge extraction methods to either fail or produce hallucinated textures that induce identity drift (see the comparison in Fig.~\ref{fig:enh_samples}). To address this, we adopt the Sauvola local adaptive binarization algorithm~\cite{sauvola2000adaptive}, a deterministic image processing technique that adaptively estimates local statistical properties for each pixel neighborhood.

For a fingerprint grayscale image $I(x,y)$, the local mean $m(x,y)$ and standard deviation $s(x,y)$ are computed within a $w \times w$ window:
\begin{equation}
  \begin{aligned}
    m(x,y) &= \frac{1}{w^2} \sum_{i=x-r}^{x+r} \sum_{j=y-r}^{y+r} I'(i,j), \\
    s(x,y) &= \sqrt{\frac{1}{w^2} \sum_{i=x-r}^{x+r} \sum_{j=y-r}^{y+r} [I'(i,j) - m(x,y)]^2},
  \end{aligned}
\end{equation}
Here, $I^{\prime}$ is the median-filtered grayscale input, $r=(w-1)/2$, and the selected window size is $w=11$. The local threshold is then:
\begin{equation}
  T(x,y) = m(x,y) \left[ 1 + k \left( \frac{s(x,y)}{R} - 1 \right) \right],
\end{equation}
where $R = 128$ is the dynamic range constant and $k = 0.007$ is the correction parameter. This mechanism automatically adapts discrimination criteria between ridge regions (high $s$) and background regions (low $s$), suppressing noise while enhancing valid texture.

The final binary image is obtained by:
\begin{equation}
  B(x,y) = \begin{cases} 0, & \text{if } (x,y) \in \mathcal{M} \text{ and } I'(x,y) > T(x,y) \\ 255, & \text{otherwise} \end{cases}
\end{equation}
Here, $\mathcal{M}$ is the binary finger-foreground mask obtained from pose estimation. A morphological opening with a $2\times2$ structuring element removes residual noise after binarization.

Although Sauvola-generated ridge edges are slightly less smooth than those from specialized extractors, its deterministic nature maximizes fidelity to the original texture, avoiding the ``blind guessing'' common in learned methods on feature-sparse regions. This aligns with the enhancement principle established in prior work~\cite{pan2024FDD,pan2025FLARE} of avoiding explicit ridge perception logic in favor of deterministic texture mapping.

\subsection{Contactless Fingerprint Modality Translation}
\label{sec:cl_gen}

Building on the Sauvola-derived image $B$, we construct a high-fidelity cross-modal translation pipeline. Following GenPrint's identity injection logic~\cite{grosz2024universal}, we combine identity binary image latent encoding with ControlNet~\cite{zhang2023adding} injection.

The translation module training is organized into three sequential {steps}:

\begin{revision}
\textbf{Step 1: Binary Image Latent Space Modeling.} A dedicated autoencoder $(E_e, D_e)$ maps a $1\times512\times512$ Sauvola image $B$ to a continuous latent $z_B=E_e(B)\in\mathbb{R}^{3\times128\times128}$, matching the size of the denoising latent. It follows the encoder--decoder layout of the VQ-VAE (channel widths 64/128/256, two residual blocks per level, two $2\times$ downsamplings) but omits vector quantization to retain fine ridge detail, and is trained with an MSE reconstruction loss. The full architecture is listed in Supplementary Table~\suppref{tab:supp_arch}.
\end{revision}

\textbf{{Step 2}: Contactless Modality Prior Learning.} A VQ-VAE ($E_{\text{cl}}$, $D_{\text{cl}}$) establishes the compressed representation space for contactless modality, followed by LDM training to capture modality-specific texture statistics. The backbone models the contactless appearance distribution.

\begin{revision}
\textbf{Step 3: Identity-Consistent Translation.} The frozen encoder $E_e$ converts the Sauvola map into $z_B$, which is fed to the ControlNet~\cite{zhang2023adding}. The branch is initialized from the encoder of the pretrained contactless denoising U-Net, and its multiscale outputs are added to the U-Net through zero-initialized convolutions, so training starts from the unmodified backbone. Only the ControlNet branch is trained; $E_e$, the contactless VQ-VAE, and the denoising U-Net remain frozen, and $D_e$ is used only in Step~1. No explicit identity loss is applied: identity is carried entirely by the spatial condition.
\end{revision}

\begin{figure}[!t]
  \centering
  \includegraphics[width=\linewidth]{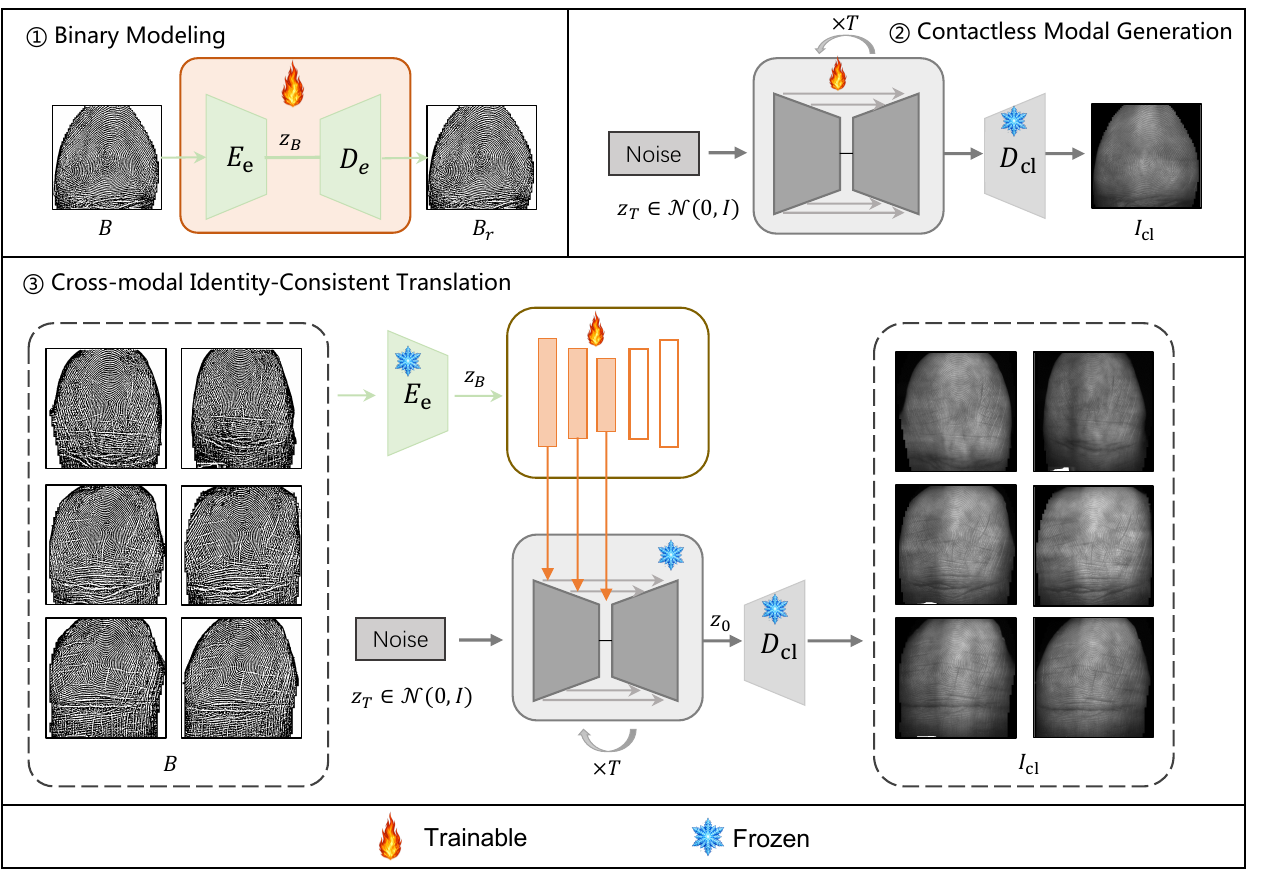}
  \caption{Training {steps} and architecture of the contactless fingerprint cross-modal translation module.}
  \label{fig:CB2CL}
\end{figure}

For training, we utilize the UWA~\cite{zhou2014benchmark} contactless fingerprint database, which provides flattened contactless images with excellent field-of-view coverage suitable for subsequent multi-pose texture mapping. All contactless images are normalized to $512 \times 512$ pixels at 500 PPI, ensuring spatial alignment with the rolled fingerprint generation module.

{
\textbf{Conditioning at Training and Inference.} Step~3 training extracts $B_{\mathrm{cl}}=\operatorname{Sauvola}(I_{\mathrm{cl}})$ from the target contactless image for pixel-aligned learning without cross-modal registration error, whereas inference feeds $B_r=\operatorname{Sauvola}(I_r)$ from the synthesized rolled identity into $E_e$. Because Sauvola binarization abstracts away modality-specific contrast while preserving ridge topology, the model generalizes directly from $B_{\mathrm{cl}}$ to $B_r$ (verified in Section~\ref{sec:identity_quant}).
}

\subsection{Multi-Pose Contactless Fingerprint Projection Simulation}
\label{sec:multi_pose}

\begin{figure*}[!t]
  \centering
  \includegraphics[width=\textwidth]{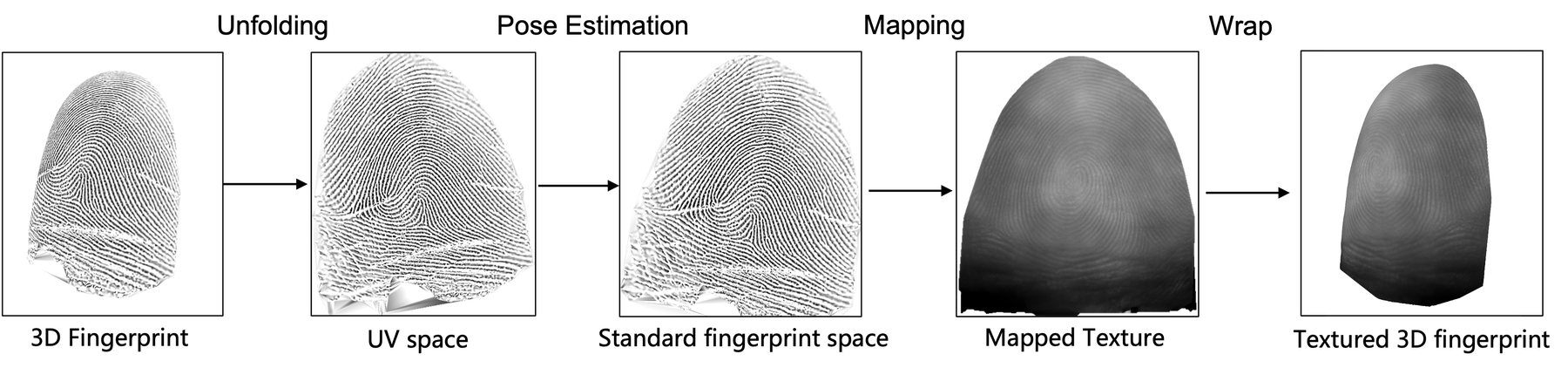}
  \caption{Illustration of the multi-pose contactless fingerprint simulation pipeline: from 3D model unfolding, pose estimation, and texture mapping to final multi-angle projection.}
  \label{fig:multi_pose_pipeline}
\end{figure*}

To render synthesized contactless textures under diverse 3D poses, we construct a simulation pipeline using finger surface point clouds as geometric carriers. The core logic establishes a mapping between 2D textures and 3D geometric manifolds to enable multi-angle projection.

We first employ the parametric unfolding algorithm proposed by Guan et al.~\cite{guan20213dpose} to map self-collected 3D fingerprint point clouds (THU Finger3D) to a 2D UV plane. Pose estimation~\cite{duan2023estimating} is then applied to rectify the unfolded fingerprint image, aligning the UV space with the standard fingerprint coordinate system. This alignment establishes point-to-point correspondence between the generated contactless texture and the 3D point cloud.

By mapping the synthesized 2D contactless texture onto the 3D finger surface, the pipeline generates contactless fingerprint images across diverse viewing poses by varying the 3D observation angle.

\begin{figure}[!t]
  \centering
  \subfloat[Front view]{\includegraphics[height=0.3\linewidth]{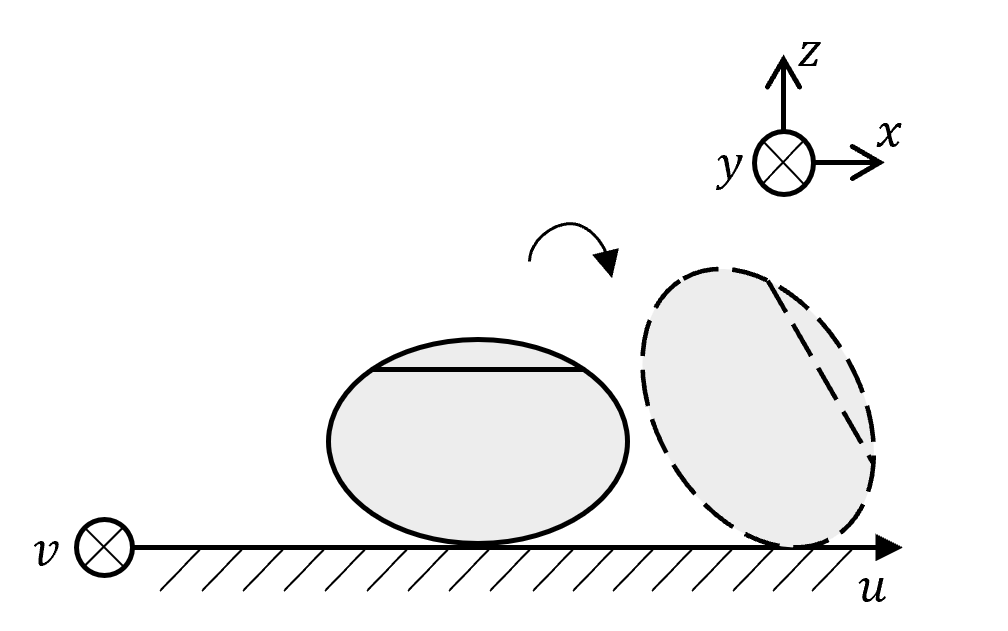}} \hfil
  \subfloat[Top view]{\includegraphics[height=0.3\linewidth]{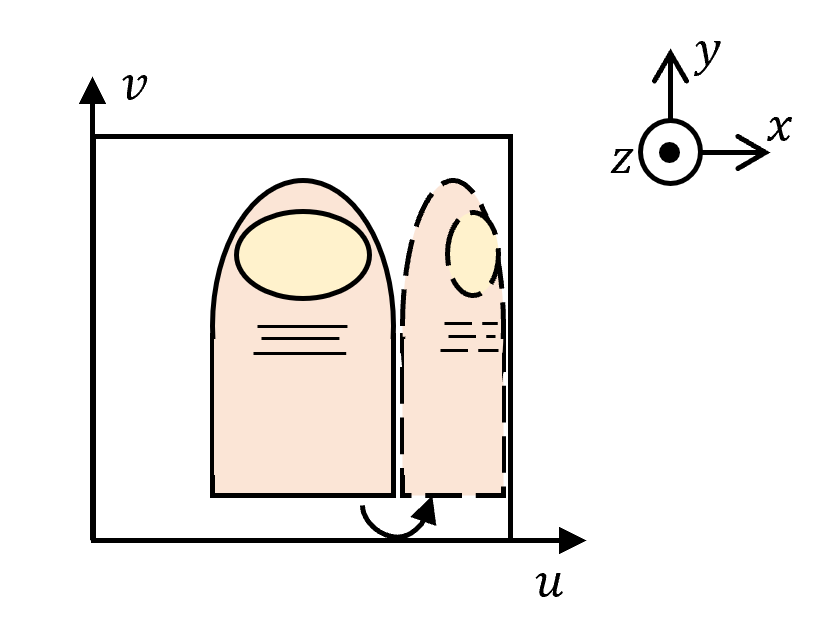}}
  \caption{Illustration of the finger roll projection mechanism under different geometric views.}
  \label{fig:uv23d}
\end{figure}

Our multi-pose simulation focuses on the finger roll angle, which is the primary challenge in contactless fingerprint recognition as it directly governs nonlinear ridge distortion~\cite{zhou2014benchmark, dong2023synthesis, tan2020towards, cui2023monocular}. We model finger rolling as a rigid-body pure rolling process on the UV plane, establishing a dynamic correspondence between 3D Euclidean space and 2D UV space, where the UV space coincides with the standard fingerprint coordinate system (see Fig.~\ref{fig:uv23d}).

During 3D model unfolding, after initial pose correction, point clouds are segmented along the principal $y$-axis with the model centroid as origin. Geodesic distances from each sampling point to the $x=0$ reference plane are computed and projected onto the $u$ coordinate axis of the 2D parameter space. Under this modeling, the $u$-$v$ axes of the UV plane maintain parallel correspondence with the $x$-$y$ axes of 3D space.

However, directly orthographically projecting the rotated 3D model onto the $x$-$y$ plane would deviate from the standard fingerprint space. To maintain spatial consistency, we introduce a displacement compensation $\Delta u$ along the $u$-axis to account for instantaneous contact point changes during finger rolling. Let $l$ be the original cross-sectional arc and $l_m$ be the arc after rotation by angle $\theta$, both unfolded to the $u$-axis using the same parameterization. The compensation is:
\begin{equation}
  \Delta u = \min(u|{l_m}^{\prime}) - \min(u|{l^{\prime}})
\end{equation}
The final projection position is:
\begin{equation}
  {l_m^{p}}^{\prime} = l_m^p + \Delta u
\end{equation}

$p$ denotes projection after rotation by $\theta$, and the prime denotes displacement compensation.

\begin{figure}[!t]
  \centering
  \subfloat[$\Delta u$ calculation principle \label{fig:delta_u}]{\includegraphics[height=0.42\linewidth]{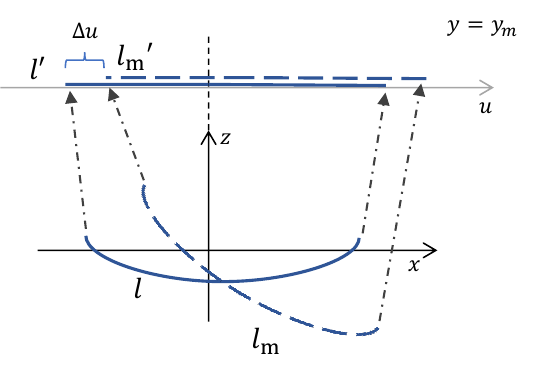}} \hfil
  \subfloat[Orthographic correction \label{fig:proj_u}]{\includegraphics[height=0.42\linewidth]{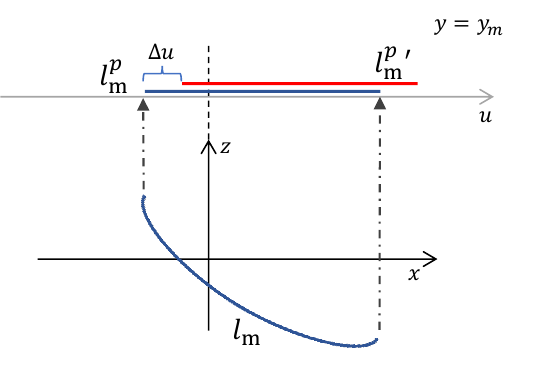}}
  \caption{Displacement compensation $\Delta u$ and orthographic projection correction, shown at an enlarged scale for readability.}
  \label{fig:delta_u_proj}
\end{figure}

Fig.~\ref{fig:delta_u_proj} illustrates the principles of $\Delta u$ calculation and the orthographic projection correction mechanism.

\begin{figure*}[!t]
  \centering
  \includegraphics[width=0.95\textwidth]{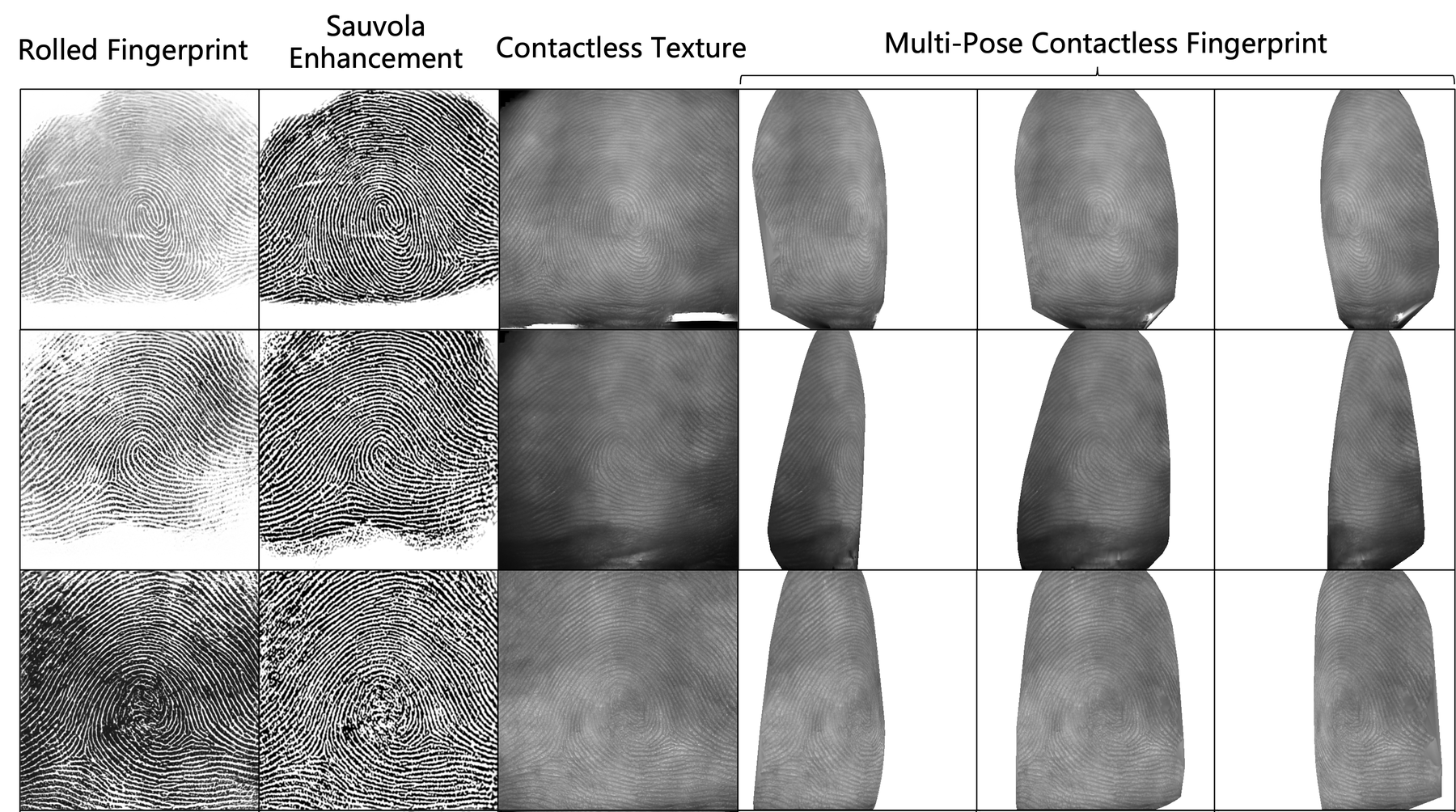}
  \caption{Representative IMPOSE identities: rolled fingerprints, binary conditions, contactless translations, and projected views.}
  \label{fig:gen_examples}
\end{figure*}

The compensation preserves source-texture correspondence, subject to resampling and visibility loss. Fig.~\ref{fig:gen_examples} presents representative samples from the full IMPOSE generation pipeline.

\subsection{Implementation Details}
\label{sec:impl}

The VQ-VAE uses base width 64, multipliers $(1,2,4)$, two residual blocks per level, and a 4,096-entry codebook, mapping $512\times512$ images to $128\times128\times3$ latents. The U-Net and ControlNet use the same width, multipliers, and block count. \rev{Supplementary Table~\suppref{tab:supp_arch} lists the full architectures of the autoencoders, the U-Net, and the ControlNet conditioning path.}

Table~\ref{tab:training_config} summarizes the training configurations for each sub-module. During the separate LDM and ControlNet training stages, a 500-step learning rate warmup is employed, where the learning rate scaling factor increases linearly from $1 \times 10^{-2}$ to $1$.

\begin{table}[!t]
\blacktable
  \caption{Training Configuration of IMPOSE Sub-Modules}
  \label{tab:training_config}
  \centering
  \begin{tabular}{lcccc}
    \toprule
    \textbf{Module} & \textbf{Batch} & \textbf{LR} & \textbf{Epochs} & \textbf{GPUs} \\
    \midrule
    Rolled VQ-VAE & 8 & $7\times10^{-5}$ & 96 & 1$\times$RTX 4090 \\
    Rolled LDM & 24 & $2\times10^{-5}$ & 780 & 1$\times$RTX 4090 \\
    \midrule
    Sauvola AE & 16 & $3.5\times10^{-4}$ & 120 & 2$\times$RTX 4090 \\
    \midrule
    CL VQ-VAE & 8 & $7\times10^{-5}$ & 60 & 1$\times$RTX 4090 \\
    CL LDM & 20 & $5\times10^{-6}$ & 310 & 1$\times$RTX 4090 \\
    CL ControlNet & 24 & $1\times10^{-6}$ & 360 & 1$\times$RTX 4090 \\
    \bottomrule
  \end{tabular}
\end{table}

\begingroup
All diffusion models are trained with AdamW, a linear noise schedule, and the $\ell_1$ loss in Eq.~(\ref{eq:ddpm_loss}). Parameter counts and generation timings are given in Supplementary Tables~\suppref{tab:supp_params} and~\suppref{tab:supp_timing}.
\endgroup

\begin{table*}[!t]
\blacktable
  \centering
  \caption{Summary of Datasets Used in IMPOSE Experiments}
  \label{tab:datasets}
  \begin{threeparttable}
    \begin{tabular}{p{0.15\textwidth}p{0.45\textwidth}p{0.25\textwidth}}
      \toprule
      \textbf{Dataset} & \textbf{Description} & \textbf{Usage} \\
      \midrule
      NIST SD302~\cite{nist302} & 2,000 rolled fingerprints, optical sensor & Identity generation training \\
      THU Rolled & 1,093 rolled fingerprints, optical sensor & Identity generation training \\
      THU Finger3D & 285 3D fingerprints, structured-light scanner & Texture mapping reference \\
      \midrule
      UWA~\cite{zhou2014benchmark}\tnote{$\dagger$}  & 1,500 fingers $\times$ 3 contactless, $\times$ 2 flattened CL, 1,500 plain & Cross-modal training\tnote{$\ddagger$} \& testing \\
      PolyU CL2CB~\cite{liu20143d} & 336 fingers $\times$ 6 contact/contactless & Testing \\
      \bottomrule
    \end{tabular}
    \begin{tablenotes}
      \item[$\dagger$] Contactless samples were pre-processed with white backgrounds and CLAHE-based contrast enhancement.
      \item[$\ddagger$] Only flattened images used for style training; no test identity leakage.
    \end{tablenotes}
  \end{threeparttable}
\end{table*}

\begin{figure*}[!t]
  \centering
  \subfloat[NIST SD302 rolled]{\includegraphics[width=0.31\textwidth]{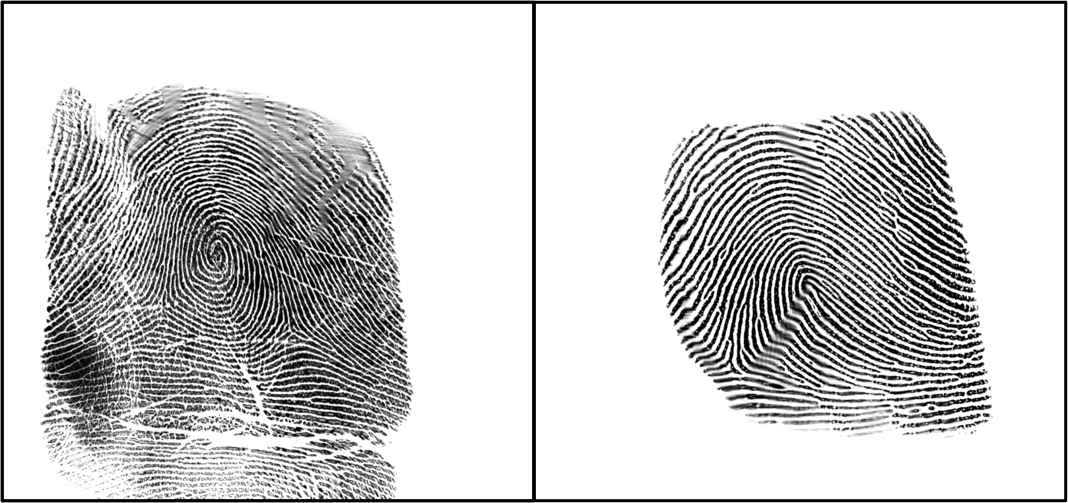}} \hfill
  \subfloat[THU Rolled]{\includegraphics[width=0.31\textwidth]{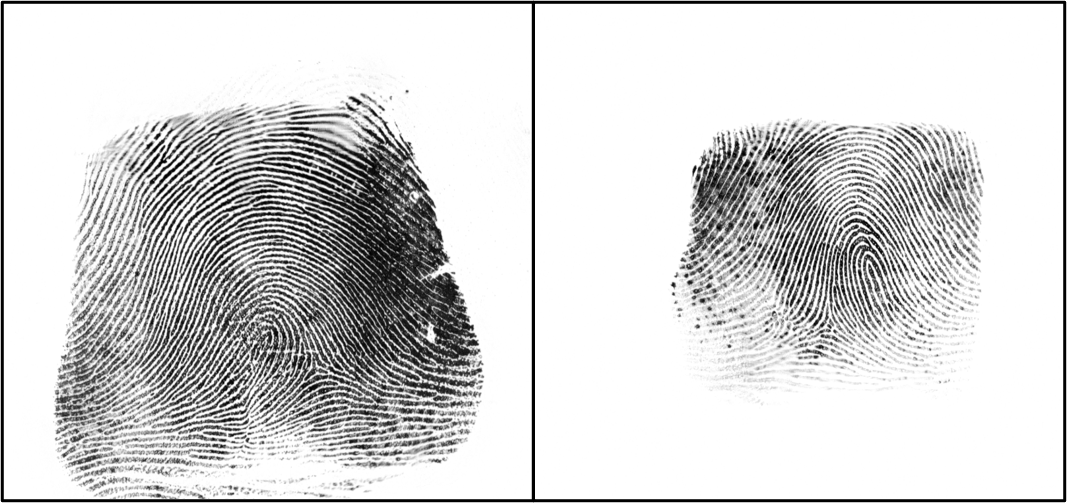}} \hfill
  \subfloat[THU Finger3D]{\includegraphics[width=0.31\textwidth]{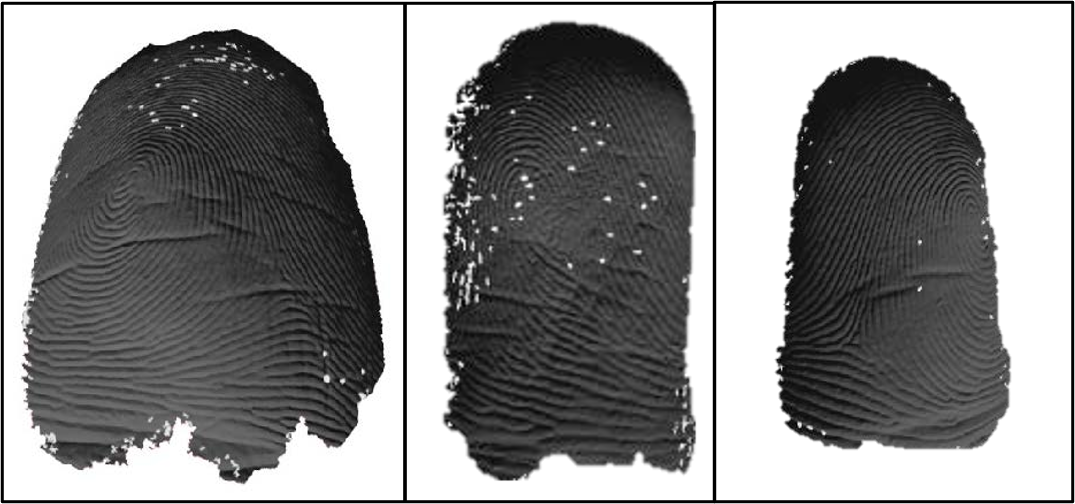}} \\
  \subfloat[UWA contactless\label{fig:uwa_cl}]{\includegraphics[width=0.31\textwidth]{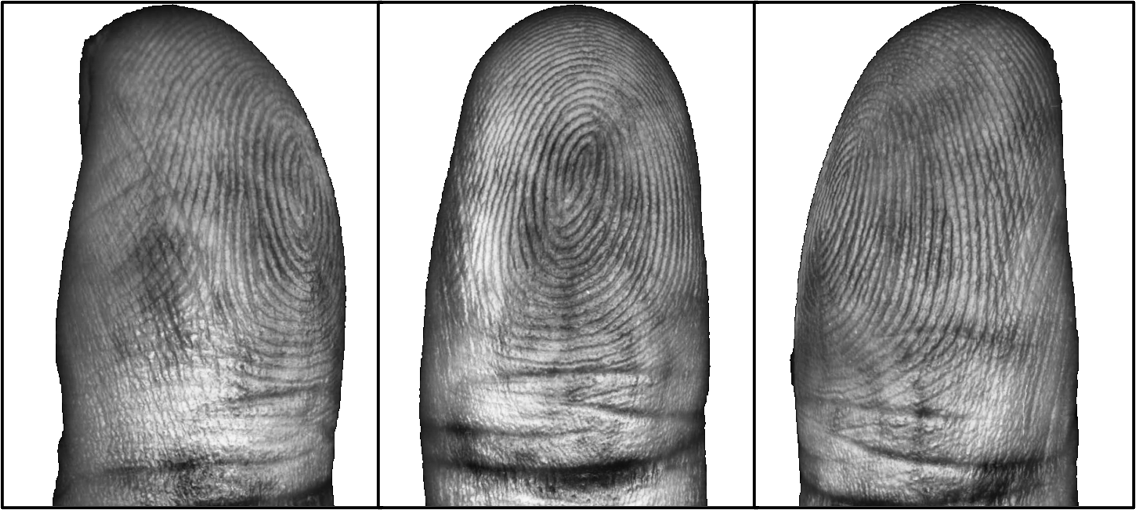}} \hfill
  \subfloat[UWA flattened CL\label{fig:uwa_fl}]{\includegraphics[width=0.31\textwidth]{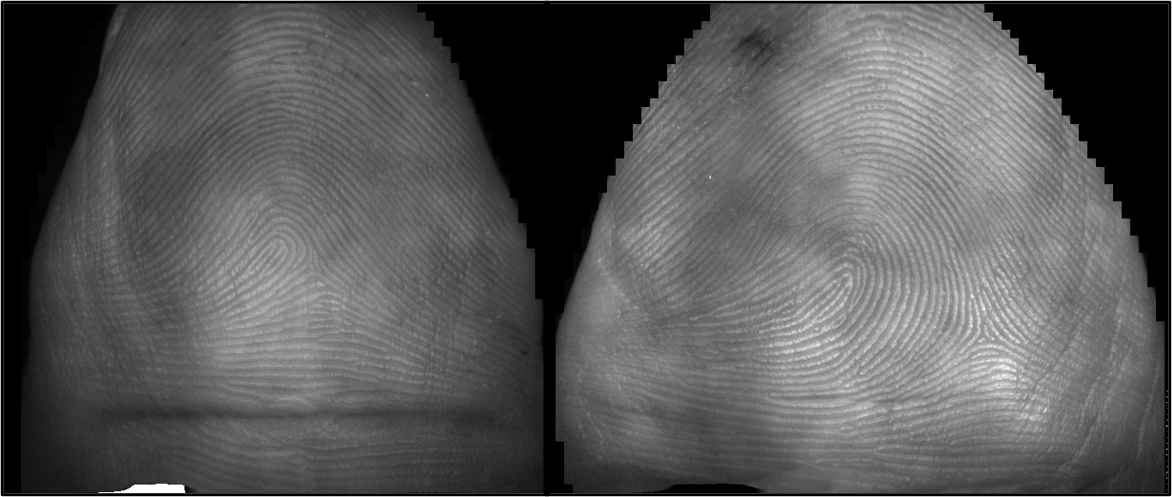}} \hfill
  \subfloat[UWA plain\label{fig:uwa_plain}]{\includegraphics[width=0.31\textwidth]{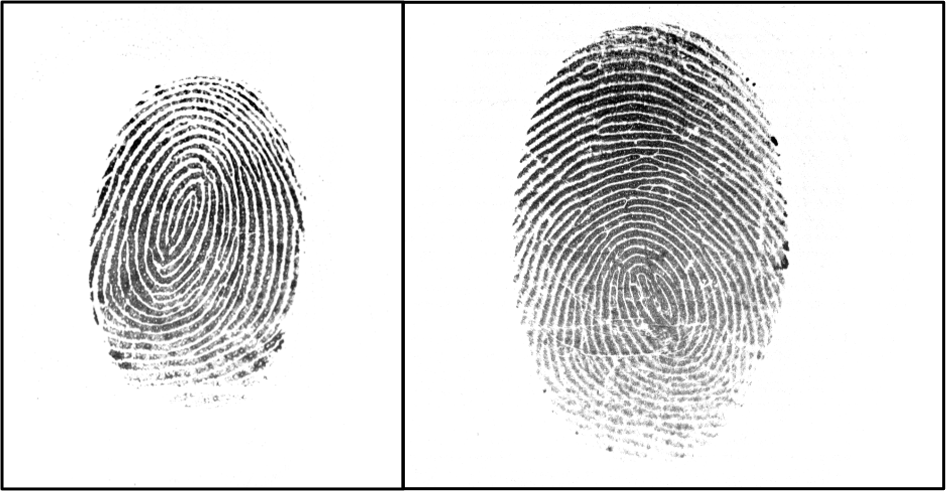}}
  \caption{Representative fingerprint samples from the datasets used in IMPOSE experiments.}
  \label{fig:dataset_samples}
\end{figure*}

During autoencoder pretraining, extensive geometric augmentation is applied (random translation within $\pm 52$ pixels, rotation within $\pm 45^\circ$, horizontal flipping) to strengthen representation robustness. For subsequent LDM and ControlNet training, only horizontal flipping is retained while translation and rotation are strictly disabled to ensure generated images maintain spatial consistency with the standard fingerprint coordinate system.

Post-generation filtering requires NFIQ 2.0~\cite{tabassi2021nist} $\geq55$ and foreground ratio $\geq0.60$. Development data are split 7:3 for tuning, then recombined for final training; evaluation identities are excluded.

\section{Experimental Datasets}

The experimental datasets encompass three key categories: high-quality rolled fingerprints for identity generation training, contactless fingerprints for style transfer training and evaluation, and 3D fingerprints for multi-pose simulation. All 2D fingerprint images are normalized to 500 PPI. Table~\ref{tab:datasets} summarizes the dataset details.

For rolled fingerprint generation, we train on 3,093 images from NIST SD302~\cite{nist302} and THU Rolled. Compared to existing methods~\cite{engelsma2022printsgan,grosz2024universal}, IMPOSE synthesizes diverse identities from a compact training set.

UWA flattened contactless images provide style training, with evaluation identities excluded. Multi-pose simulation samples templates from 285 THU Finger3D point clouds.

For evaluation, we adopt the protocol established by Cui et al.~\cite{cui2023monocular}. For the UWA dataset~\cite{zhou2014benchmark}, the gallery is formed by the initial contact-based plain fingerprints of the first 1,000 fingers, while the query set consists of the corresponding contactless images captured at three angles. For the PolyU CL2CB dataset~\cite{liu20143d}, contactless fingerprints are pre-processed by scaling them to the mean ridge period of the contact-based samples, notably omitting any geometric warping.

\section{Experimental Results}

\subsection{Rolled Fingerprint Generation Evaluation}

To quantitatively verify the rolled fingerprint generation module's ability to capture the target domain distribution, we randomly sample 3,000 synthetic rolled fingerprints (comparable to the 3,093 training samples).

\begin{figure}[!t]
  \centering
  \includegraphics[width=\linewidth]{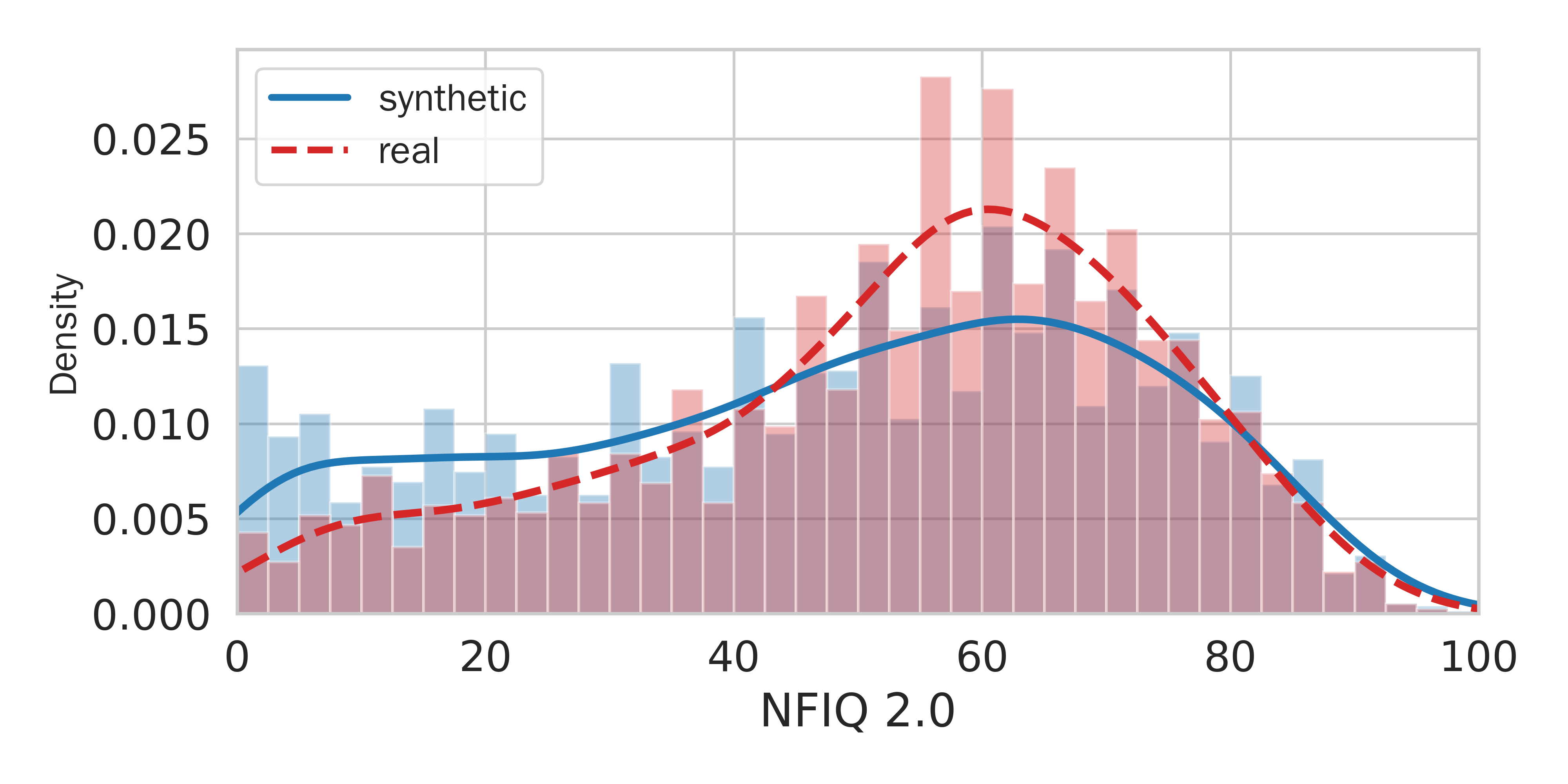}
  \caption{NFIQ 2.0 quality distribution comparison between synthetic rolled fingerprints and real training samples.}
  \label{fig:nfiq}
\end{figure}

\textbf{Image Quality Distribution.} We employ NFIQ 2.0~\cite{tabassi2021nist} to analyze quality distribution alignment between synthetic and real samples. As shown in Fig.~\ref{fig:nfiq}, the synthetic distribution has a shape similar to the training set, with mean NFIQ 2.0 scores of 47.59 (synthetic) vs.\ 52.81 (real) and standard deviations of 24.51 vs.\ 20.94. The similar peaks and coverage intervals indicate that IMPOSE captures key aspects of the rolled-fingerprint quality distribution.

\begin{figure}[!t]
  \centering
  \includegraphics[width=\linewidth]{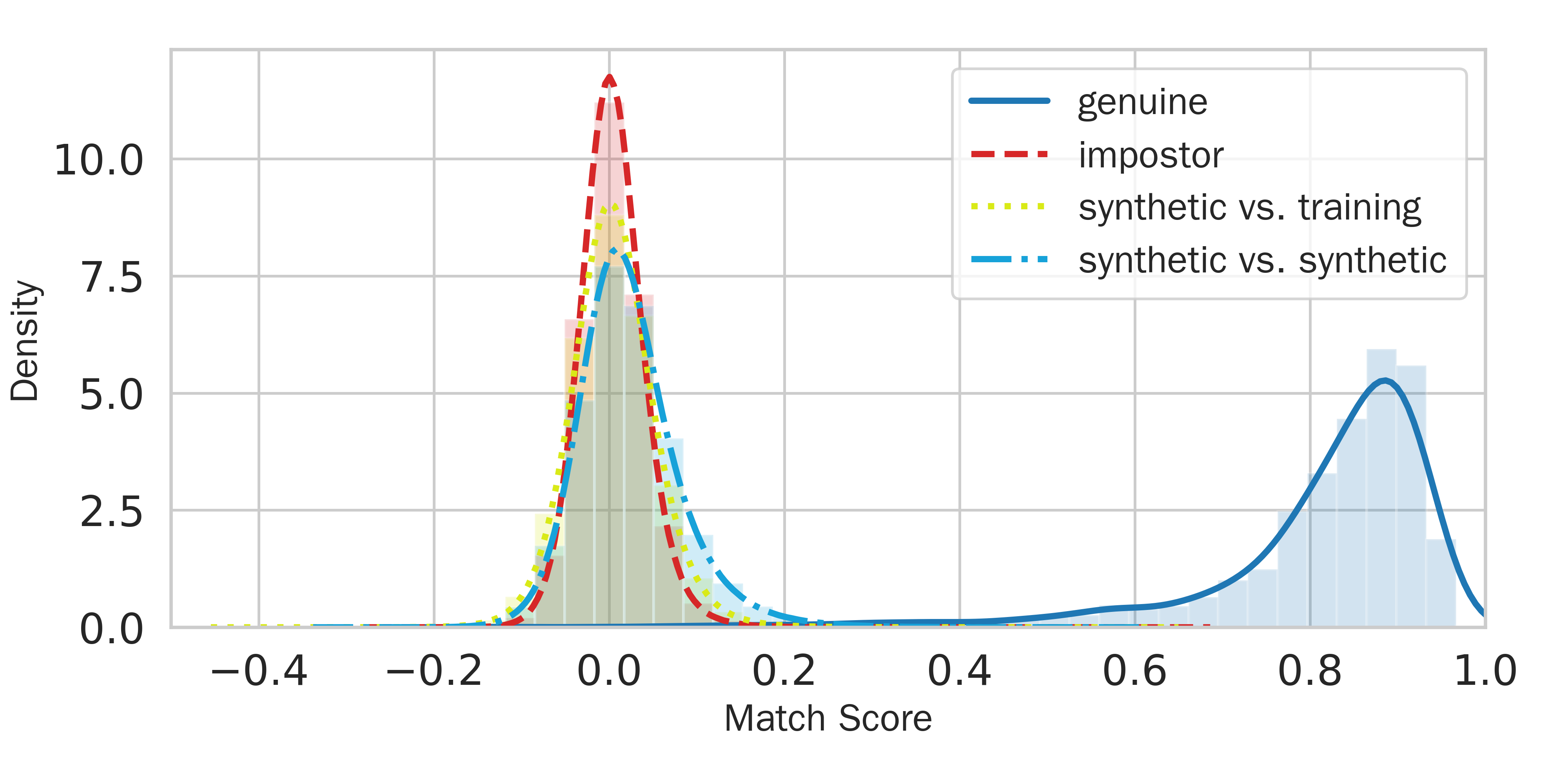}
  \caption{Match score distributions evaluating identity consistency, uniqueness, and privacy preservation of synthetic rolled fingerprints.}
  \label{fig:match_score}
\end{figure}

\begin{figure*}[!t]
  \centering
  \includegraphics[width=0.84\textwidth]{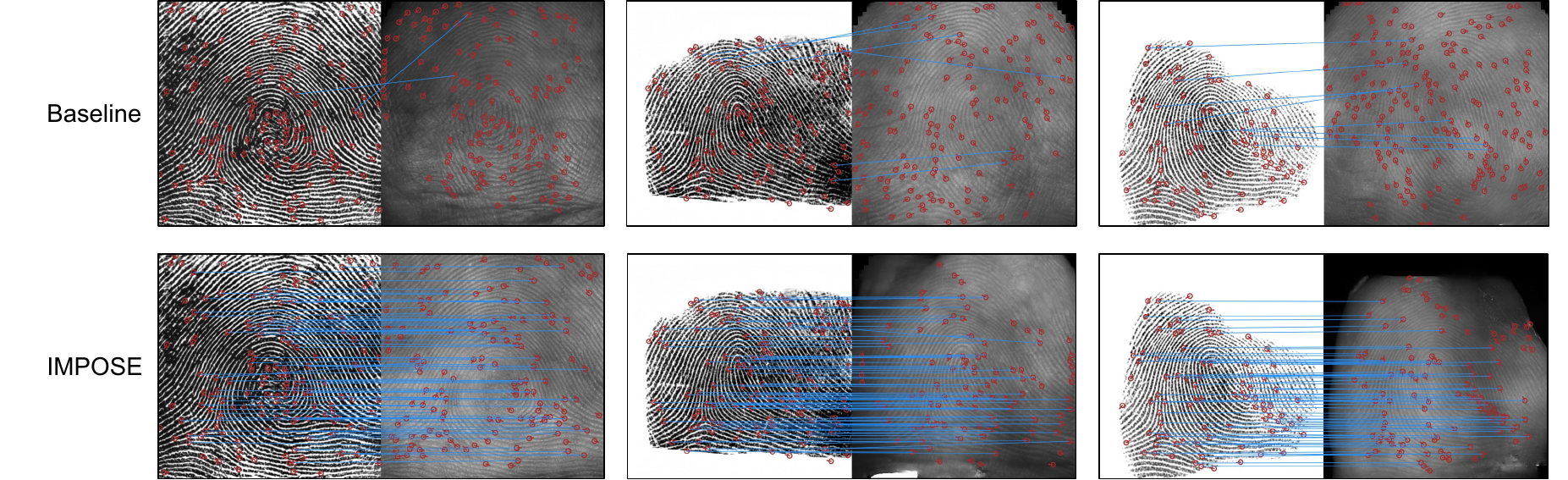}
  \caption{Identity control using a rigidly aligned contact-image baseline and IMPOSE with target-derived Sauvola maps for training and rolled-derived maps for generation.}
  \label{fig:id_control}
\end{figure*}

\textbf{Identity Uniqueness and Privacy.} Fig.~\ref{fig:match_score} compares FDD~\cite{pan2024FDD} score distributions for real genuine/impostor pairs, synthetic-vs.-training pairs, and inter-identity synthetic pairs. Both synthetic-vs.-training and synthetic-vs.-synthetic distributions closely match the real impostor curve, confirming that the generator produces distinct identities without memorizing training samples.

{
\subsection{Quality Filtering and Retained-Identity Diversity}
\label{sec:filter}
Table~\ref{tab:filter} gives sequential filtering statistics for all 10,008 sampled rolled identities. NFIQ 2.0 ($\geq55$) retains 4,731 samples, and foreground filtering ($\text{mask}\geq0.60$, evaluated on the NFIQ-passing subset) retains 1,712 identities (17.11\%) for downstream synthesis.
\begin{table}[!t]
\blacktable
  
  \centering
  \small
  \setlength{\tabcolsep}{3pt}
  \caption{Sequential Quality Filtering}
  \label{tab:filter}
  \resizebox{0.95\linewidth}{!}{
  \begin{tabular}{lrrr}
    \toprule
    Criterion & Count & Denominator & Rate (\%) \\
    \midrule
    NFIQ $\geq55$ & 4,731 & 10,008 & 47.27 \\
    NFIQ $<55$ & 5,277 & 10,008 & 52.73 \\
    Mask $\geq0.60$, given NFIQ pass & 1,712 & 4,731 & 36.19 \\
    Mask $<0.60$, given NFIQ pass & 3,019 & 4,731 & 63.81 \\
    Both criteria pass & 1,712 & 10,008 & 17.11 \\
    \bottomrule
  \end{tabular}}
\end{table}
Filtering is applied post hoc so that the generator models the full rolled-image distribution during training while supplying clean, high-coverage references for translation and 3D projection. As shown in Fig.~\ref{fig:tsne}, the 1,712 retained identities remain well dispersed across the FDD feature space and across NFIQ 2.0 levels above the threshold.
\begin{figure}[!t]
\centering\includegraphics[width=.72\linewidth]{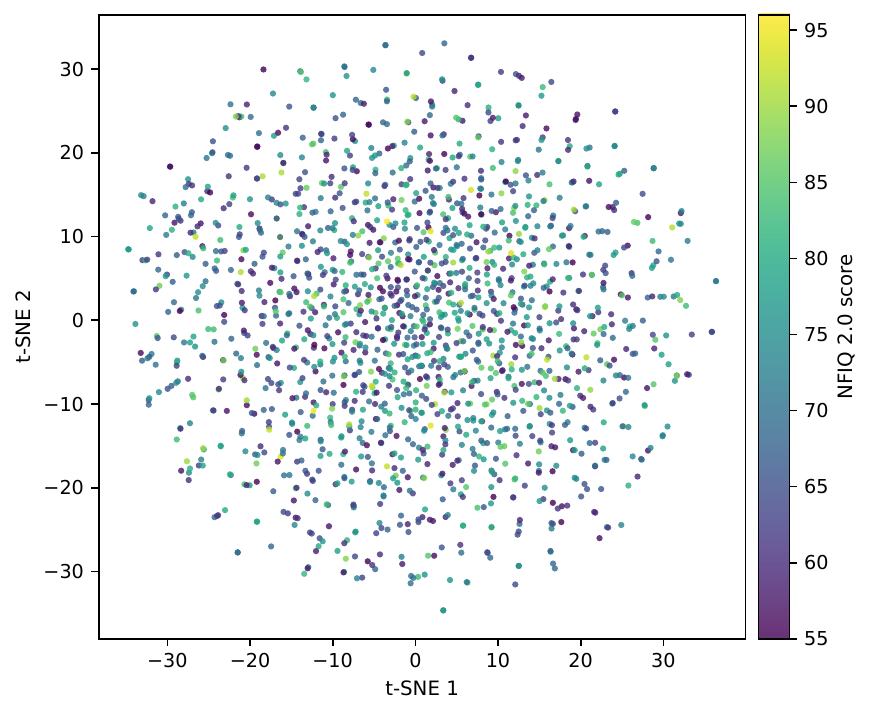}
\caption{FDD features of 1,712 retained identities visualized by PCA and t-SNE; colors denote NFIQ scores.}\label{fig:tsne}
\end{figure}
}

\subsection{Contactless Fingerprint Generation Evaluation}
\label{sec:cl_gen_eval}

We next evaluate the structural fidelity and cross-dataset transferability of the translation module.

\textbf{Identity Consistency Analysis.} We compare Sauvola-guided conditioning against a baseline that rigidly aligns paired contact and contactless impressions via VeriFinger~\cite{nist2020verifinger} minutiae and conditions translation on the aligned contact image. Because rigid alignment cannot remove nonlinear distortion between paired impressions, the baseline exhibits local ridge drift, whereas conditioning on the pixel-aligned Sauvola map preserves fine ridge correspondence (Fig.~\ref{fig:id_control}).

\begin{figure}[!t]
  \centering
  \includegraphics[width=.90\linewidth]{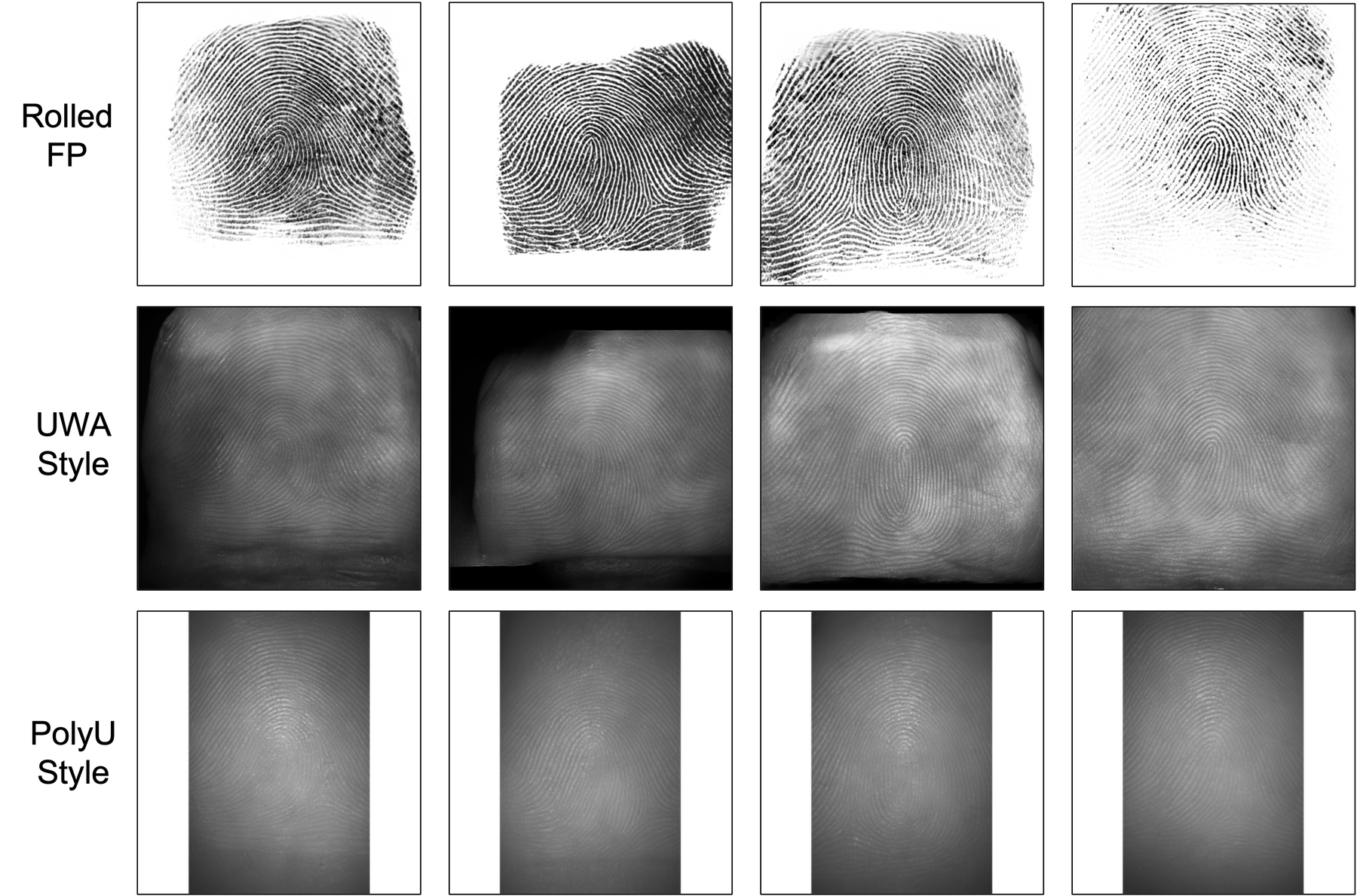}
  \caption{Generation results of IMPOSE under different contactless fingerprint styles (UWA and PolyU CL2CB).}
  \label{fig:polyu_gen}
\end{figure}

\textbf{Cross-Database Modality Generalization.} Fig.~\ref{fig:polyu_gen} shows transfer results on PolyU CL2CB. The generated samples capture its characteristic localized distribution and low-contrast visual style while preserving ridge-level correspondence, indicating that the translation model can transfer to a contactless modality not used for style training.

\subsection{Identity Preservation after 3D Projection}
\label{sec:identity_quant}
\begin{revision}
Before projection, DMD~\cite{pan2024DMD} and VeriFinger~\cite{nist2020verifinger} both reach 100\% Rank-1 on the 1,712 Stage~2 translations (Supplementary Table~\suppref{tab:supp_stage2_identity}). To test whether identity survives projection, we use all 15,136 final Stage~3 images as queries against the 1,712 rolled references, extracting minutiae from each projected image.\footnote{\rev{The schedule requests $1{,}712\times9=15{,}408$ views, but the eight non-frontal angles are drawn with replacement and files are named by identity and angle, so a repeated draw overwrites an existing view. All 272 missing views result from such repeated draws, and no projection failed: 1,450 identities have nine views, 252 have eight, and 10 have seven (Supplementary Table~\suppref{tab:supp_projection_counts}).}}

\begin{table}[!t]
\blacktable
\revisionfloatcolor
\centering\small\setlength{\tabcolsep}{3pt}
\caption{\rev{Identity Consistency of Final Stage 3 Projections (\%)}}
\label{tab:identity_quant}
\begin{tabular}{lrrrrr}
\toprule
\multirow{2}{*}{Roll ($^\circ$)}&\multirow{2}{*}{$N$}&\multicolumn{2}{c}{DMD}&\multicolumn{2}{c}{VeriFinger}\\
\cmidrule(lr){3-4}\cmidrule(lr){5-6}
&&R-1&EER&R-1&EER\\
\midrule
All&15,136&98.57&0.839&98.64&0.805\\
$[-59,-31]$&3,378&97.28&1.362&97.42&1.498\\
$[-30,-10]$&3,339&99.88&0.180&99.52&0.407\\
$0$&1,712&99.88&0.116&99.77&0.192\\
$[10,30]$&3,340&99.91&0.119&99.73&0.206\\
$[31,59]$&3,367&96.55&1.634&97.33&1.190\\
\bottomrule
\end{tabular}
\end{table}

Table~\ref{tab:identity_quant} reports results by roll-angle group. 
Within $\pm30^\circ$, Rank-1 stays above 99.5\% for both matchers. Beyond $\pm30^\circ$ it drops to 96.55--97.42\%. At a single threshold giving an overall FAR $\leq0.1\%$ (TAR 98.26\% for DMD and 98.50\% for VeriFinger), the FNMR of the outer groups rises to 2.9--3.8\%, while per-group FARs range from 0.05\% to 0.16\%. The drop follows from the projection geometry: at large roll, part of the ridge area turns away from the camera and the remainder near the silhouette is foreshortened, leaving fewer reliable minutiae. In the extreme case (roll $-58^\circ$), only one minutia is extracted, below the five required by DMD, so the query cannot be scored. Full operating points and angle-stratified DET curves are given in Supplementary Tables~\suppref{tab:supp_projection_operating}--\suppref{tab:supp_projection_roc} and Figs.~\suppref{fig:supp_projection_dmd}--\suppref{fig:supp_projection_vf}.
\end{revision}

{
\subsection{Sensitivity of Sauvola Conditioning}
\label{sec:sauvola_sensitivity}
We evaluate $w\in\{7,11,15,21\}$ and $k\in\{0.003,0.007,0.02,0.05\}$ ($R=128$) on normal, low-contrast, and locally noisy samples. Across these examples, $k$ has the stronger visible effect: $0.003$--$0.007$ preserve similar ridge structure, whereas larger values increasingly merge ridges. We use $w=11$, $k=0.007$; the complete parameter grids are provided in {Supplementary Fig.~\suppref{fig:supp_sauvola}} and are intended as a qualitative sensitivity analysis.
}

\subsection{FDD Performance Enhancement with IMPOSE Data}
\label{sec:fdd_enhance}

We evaluate the impact of IMPOSE-synthesized data on FDD~\cite{pan2024FDD}, which consists of a fingerprint pose estimation module~\cite{guan2025finger} and a dense descriptor extraction module. Rank-1 identification rate, TAR@FAR=0.1\%, and Equal Error Rate (EER) serve as evaluation metrics.

{\textbf{Experimental Setup.} Of 10,008 sampled identities, 1,712 pass filtering; each contributes one frontal view and up to eight roll-angle views in $[-59^\circ,-10^\circ]\cup[10^\circ,59^\circ]$, giving the 15,136 projected images described in Section~\ref{sec:identity_quant}.}

The baseline uses models trained solely on real data. Fine-tuning is performed on the pose estimation module (3 epochs) and descriptor module (5 epochs) separately and jointly. \rev{Unlike the controlled comparison in Section~\ref{sec:stage3_ablation}, the fine-tuned pose and joint settings also fine-tune the pose estimator.}

\begin{table*}[!t]
\blacktable
  \centering
  \caption{Cross-Modal Matching Performance of FDD on UWA Database (\%)}
  \label{tab:fdd_uwa}
  \small
  \resizebox{0.90\textwidth}{!}{
  \begin{threeparttable}
    \begin{tabular}{lccccccccc}
      \toprule
      \multirow{2}{*}{\textbf{Method}} & \multicolumn{3}{c}{\textbf{Left}} & \multicolumn{3}{c}{\textbf{Front}} & \multicolumn{3}{c}{\textbf{Right}} \\
      \cmidrule(lr){2-4} \cmidrule(lr){5-7} \cmidrule(lr){8-10}
      & \textbf{R-1} & \textbf{TAR} & \textbf{EER} & \textbf{R-1} & \textbf{TAR} & \textbf{EER} & \textbf{R-1} & \textbf{TAR} & \textbf{EER} \\
      \midrule
      Cui et al.~\cite{cui2023monocular} & -- & -- & 16.01 & -- & -- & 3.59 & -- & -- & 16.47 \\
      \midrule
      FDD~\cite{pan2024FDD} baseline & 60.34 & 59.94 & 15.06 & 95.78 & 95.58 & 2.19 & 63.86 & 63.25 & 17.17 \\
      FT Descriptor & 59.54 & 58.23 & 16.87 & 96.99 & 96.89 & 1.71 & 67.07 & 65.26 & 16.57 \\
      FT Pose & 68.88 & 68.88 & 11.67 & 95.68 & 94.78 & 2.81 & 71.89 & 72.79 & 12.85 \\
      \rowcolor{gray!15}
      FT Pose $+$ Descriptor & 69.96 & 69.88 & \textbf{10.94} & 96.79 & 96.18 & 2.21 & 73.49 & 73.49 & \textbf{12.63} \\
      \bottomrule
    \end{tabular}
    \begin{tablenotes}[flushleft]
\setlength{\labelsep}{0pt}
      \raggedright
      \item[] TAR = TAR@FAR=0.1\%. R-1 = Rank-1. FT = Fine-Tuned.
    \end{tablenotes}
  \end{threeparttable}}
\end{table*}

\begin{figure}[!t]
  \centering
  \includegraphics[width=\linewidth]{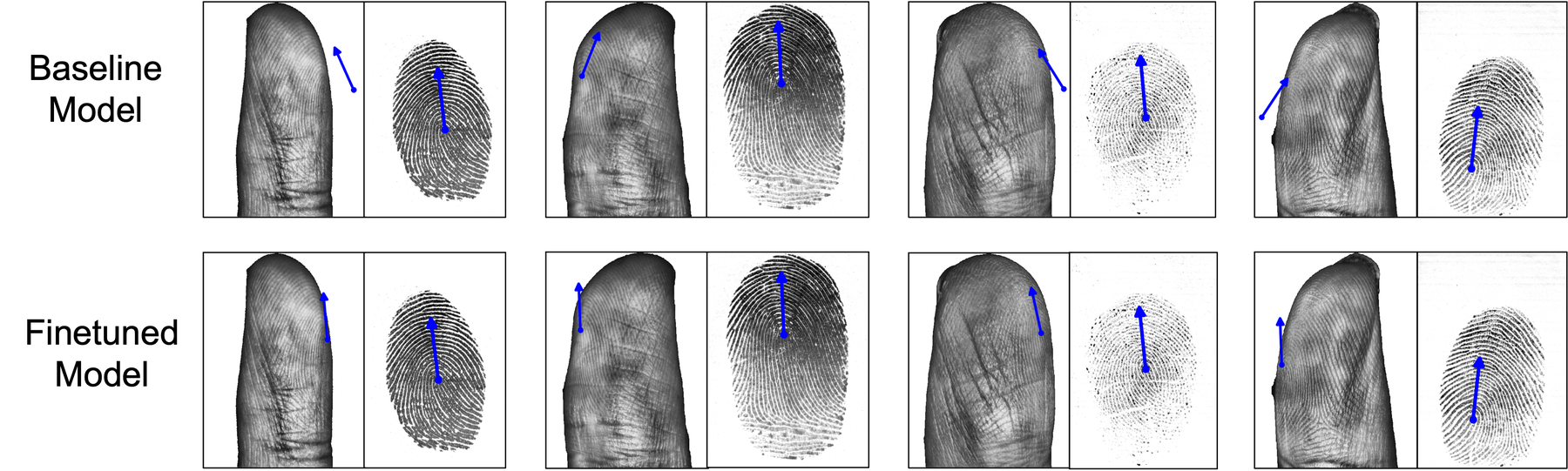}
  \caption{Visualization of pose estimation results before and after fine-tuning with IMPOSE data on large-angle contactless fingerprints.}
  \label{fig:pose_aug}
\end{figure}

\textbf{Quantitative Results.} Table~\ref{tab:fdd_uwa} reports results on UWA partitioned into left, front, and right roll subsets. The baseline FDD degrades markedly under large roll angles, with the pose estimation module exhibiting substantial geometric bias on non-central regions with severe nonlinear distortion. Fine-tuning the pose module with IMPOSE data improves prediction accuracy at extreme angles, reducing EER from 15.06\% to 11.67\% (left) and from 17.17\% to 12.85\% (right). Fine-tuning the descriptor module benefits the front subset (EER from 2.19\% to 1.71\%), reflecting enhanced modality-specific feature extraction. Joint fine-tuning achieves the optimal overall results. As shown in Fig.~\ref{fig:pose_aug}, fine-tuning the pose estimation module with IMPOSE data improves geometric prediction accuracy for large-angle contactless fingerprints while preserving robust estimation on standard contact-based fingerprints.

\begin{figure}[!t]
  \centering
  \subfloat[CMC curves]{\includegraphics[height=0.4\linewidth]{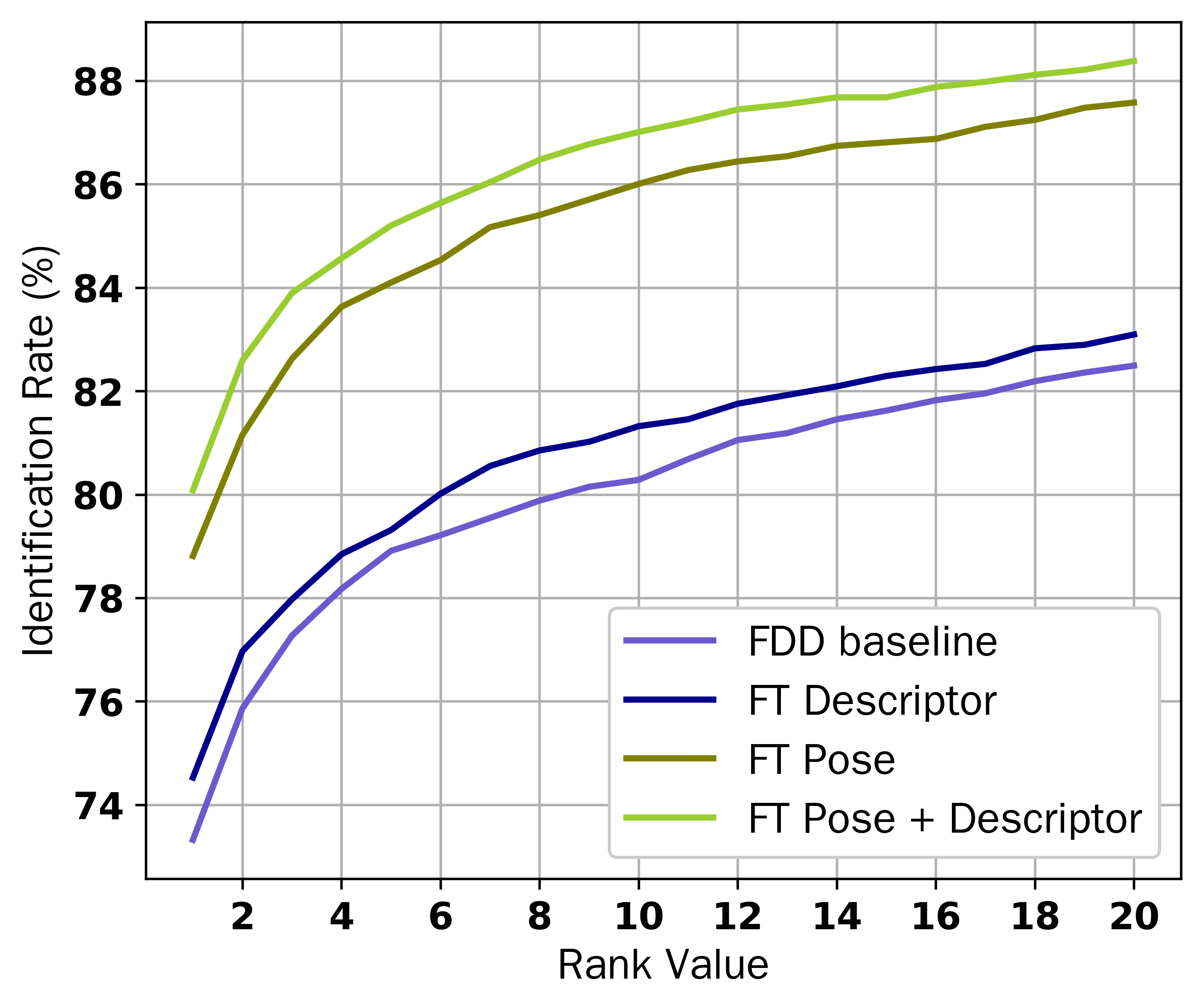}} \hfil
  \subfloat[DET curves]{\includegraphics[height=0.4\linewidth]{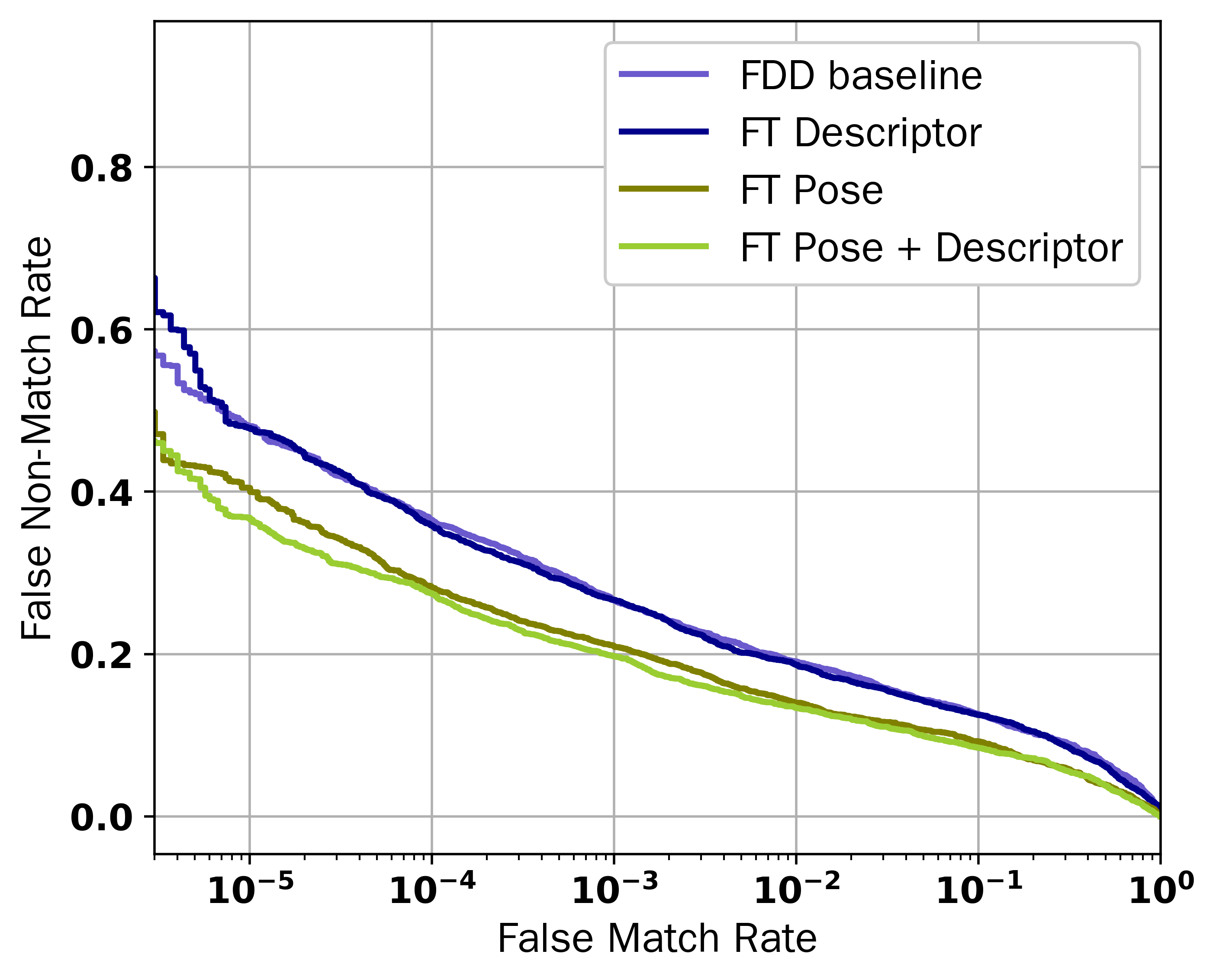}}
  \caption{CMC and DET curves of FDD under different fine-tuning strategies on the UWA database (all subsets combined).}
  \label{fig:uwa_cmc_det}
\end{figure}

Fig.~\ref{fig:uwa_cmc_det} presents CMC and DET curves for all fine-tuning strategies, with joint fine-tuning performing best on the combined evaluation despite tradeoffs across individual subsets.

Compared with the monocular depth-based unwarping method of Cui et al.~\cite{cui2023monocular}, FDD fine-tuned with IMPOSE data achieves a lower EER without explicit geometric preprocessing, indicating that physically aligned synthetic data can improve implicit modeling of nonlinear distortions.

\subsection{Controlled Comparison of 3D Projection and 2D Augmentation}
\label{sec:stage3_ablation}
\begin{revision}
To isolate the contribution of 3D projection, we fine-tune FDD under three settings on the same 1,712 identities: (A) Stage~2 frontal images with standard affine and noise augmentation; (B) A plus PCA-based thin-plate-spline (PCA-TPS) deformation, a conventional 2D model of nonlinear distortion; and (C) Stage~3 projected views with the same augmentation as B. All runs start from the same FDD weights and classification heads, keep the pose estimator fixed, train the descriptor and heads with identical AdamW settings and sample order, and stop after 210 updates with an effective batch of 48 pairs. The final checkpoint is evaluated without test-set model selection (Supplementary Tables~\suppref{tab:supp_ablation_sources}--\suppref{tab:supp_ablation_protocol}). The initial FDD is reevaluated under the same protocol as the fine-tuned models.

\begin{table}[!t]
\blacktable
\revisionfloatcolor
\centering\small\setlength{\tabcolsep}{3pt}
\caption{\rev{Controlled Descriptor Fine-Tuning with Different Training Views (\%)}}
\label{tab:stage3_ablation}
\begin{threeparttable}
\begin{tabular}{llrrrr}
\toprule
Test&Setting&R-1&TAR$_{.1}$&TAR$_{.01}$&EER\\
\midrule
UWA&Initial FDD&73.33&73.39&63.49&12.01\\
&A&69.95&69.75&59.87&12.92\\
&B&70.15&70.82&61.04&12.55\\
&C&71.02&70.55&60.58&11.95\\
\midrule
PolyU&Initial FDD&91.82&88.17&79.82&3.19\\
&A&90.53&85.42&75.98&3.93\\
&B&89.14&84.42&74.84&4.26\\
&C&93.25&88.38&80.65&3.40\\
\bottomrule
\end{tabular}
\begin{tablenotes}[flushleft]
\setlength{\labelsep}{0pt}
\footnotesize\raggedright
\item[] R-1 = Rank-1. TAR$_{.1}$ and TAR$_{.01}$ denote TAR at FAR=0.1\% and 0.01\%, respectively.
\end{tablenotes}
\end{threeparttable}
\end{table}

Table~\ref{tab:stage3_ablation} shows that C achieves the highest Rank-1 and lowest EER among the three settings on both databases. Compared with the 2D PCA-TPS baseline (B), C improves Rank-1 by 0.87 and 4.12 points and reduces EER by 0.60 and 0.86 points on UWA and PolyU, respectively, while also improving low-FAR TARs on PolyU. Differences quoted in the text are computed from unrounded values. Adding 2D TPS warping to frontal images (B vs.\ A) does not improve PolyU results, confirming that generic 2D warping cannot replace 3D projection. While A and B generate 2D-warped views on the fly from the 1,712 frontal images across the 10,080 pair presentations, C draws from the 15,136 pre-projected 3D views under the same online augmentation, providing realistic perspective foreshortening and visibility changes.

Compared with the initial FDD, C improves Rank-1 (93.25\% vs.\ 91.82\%) and both low-FAR TARs on PolyU and achieves a slightly lower EER on UWA (11.95\% vs.\ 12.01\%), whereas A and B fall below the initial model. On UWA, fixing the pose estimator and training with newly initialized heads for 210 updates lowers Rank-1 and TAR, consistent with Table~\ref{tab:fdd_uwa} where large-angle gains on UWA come mainly from fine-tuning the pose estimator. Thus, while this controlled comparison verifies the advantage of 3D projection under a fixed budget, the full benefit of IMPOSE is obtained through joint pose and descriptor fine-tuning (Section~\ref{sec:fdd_enhance}).
\end{revision}

\subsection{Comprehensive Comparison with State-of-the-Art}

Table~\ref{tab:comparison} compares the jointly fine-tuned FDD against recent state-of-the-art methods including Cui et al.~\cite{cui2023monocular}, Jia et al.~\cite{jia2024improving}, and Dong et al.~\cite{dong2025bridging}, all of which follow the ``explicit geometric correction'' paradigm. Performance metrics are cited from their original papers for protocols consistent with ours.

\begin{table}[!t]
\blacktable
  \centering
  \caption{Cross-Modal Matching Comparison on UWA and PolyU CL2CB (\%)}
  \label{tab:comparison}
  \begin{threeparttable}
    \begin{tabular}{lcccccc}
      \toprule
      \multirow{2}{*}{\textbf{Method}} & \multicolumn{3}{c}{\textbf{UWA}} & \multicolumn{3}{c}{\textbf{PolyU CL2CB}} \\
      \cmidrule{2-4} \cmidrule{5-7}
      & \textbf{R-1} & \textbf{TAR} & \textbf{EER} & \textbf{R-1} & \textbf{TAR} & \textbf{EER} \\
      \midrule
      Cui et al.~\cite{cui2023monocular} & -- & -- & 12.72 & -- & -- & 5.00 \\
      Jia et al.~\cite{jia2024improving} & -- & -- & 11.16 & -- & -- & --  \\
      Dong et al.~\cite{dong2025bridging} & -- & -- & 9.93 & -- & -- & 5.86 \\
      \midrule
      FDD~\cite{pan2024FDD} & 73.33 & 73.39 & 12.01 & 91.82 & 88.17 & 3.20 \\
      Fine-tuned FDD & 80.09 & 80.29 & \textbf{8.74} & 95.68 & 92.37 & \textbf{2.26} \\
      \bottomrule
    \end{tabular}
    \begin{tablenotes}[flushleft]
\setlength{\labelsep}{0pt}
      \raggedright
      \item[] TAR = TAR@FAR=0.1\%.
    \end{tablenotes}
  \end{threeparttable}
\end{table}

On UWA (all poses combined), the fine-tuned FDD achieves an EER of 8.74\%, compared with 9.93\% for Dong et al.~\cite{dong2025bridging}, 11.16\% for Jia et al.~\cite{jia2024improving}, and 12.72\% for Cui et al.~\cite{cui2023monocular}. On PolyU CL2CB, it reaches 95.68\% Rank-1 identification and 2.26\% EER. Under the reported protocols, these results support synthetic-data fine-tuning as an alternative to explicit geometric preprocessing.

\subsection{Generalizability to Mainstream Fixed-Length Representations}
\label{sec:generalize}

To assess transfer across recognition architectures, we evaluate DeepPrint~\cite{engelsma2021deepprint} and AFRNet~\cite{grosz2023afrnet}, which rely on Spatial Transformer Networks for implicit spatial normalization rather than explicit pose correction.

\begin{table*}[!t]
\blacktable
  \centering
  \caption{Cross-Modal Matching of Mainstream Fixed-Length Representations with IMPOSE Data (\%)}
  \label{tab:other_multi_pose}
  \small
  \resizebox{0.90\textwidth}{!}{
  \begin{threeparttable}
    \begin{tabular}{lccccccccc}
      \toprule
      \multirow{2}{*}{\textbf{Method}} & \multicolumn{3}{c}{\textbf{Left}} & \multicolumn{3}{c}{\textbf{Front}} & \multicolumn{3}{c}{\textbf{Right}} \\
      \cmidrule(lr){2-4} \cmidrule(lr){5-7} \cmidrule(lr){8-10}
      & \textbf{R-1} & \textbf{TAR} & \textbf{EER} & \textbf{R-1} & \textbf{TAR} & \textbf{EER} & \textbf{R-1} & \textbf{TAR} & \textbf{EER} \\
      \midrule
      DeepPrint~\cite{engelsma2021deepprint} & 29.62 & 27.31 & 18.05 & 75.60 & 67.67 & 4.82 & 34.94 & 32.93 & 13.86 \\
      FT DeepPrint & 51.20 & 48.69 & 13.55 & 86.35 & 85.94 & 4.52 & 55.72 & 54.72 & 12.85 \\
      \midrule
      AFRNet~\cite{grosz2023afrnet} & 41.47 & 41.06 & 20.68 & 91.77 & 90.46 & 4.11 & 51.91 & 51.81 & 15.86 \\
      FT AFRNet & 55.91 & 53.92 & 14.66 & 91.27 & 90.66 & 3.82 & 54.22 & 54.52 & 14.67 \\
      \bottomrule
    \end{tabular}
    \begin{tablenotes}[flushleft]
\setlength{\labelsep}{0pt}
      \raggedright
      \item[] TAR = TAR@FAR=0.1\%. FT = Fine-Tuned.
    \end{tablenotes}
  \end{threeparttable}}
\end{table*}

Table~\ref{tab:other_multi_pose} shows that fine-tuning with IMPOSE data reduces EER across all pose subsets for both networks. DeepPrint achieves substantial Rank-1 gains of 21.58 and 20.78 points on the left and right subsets, while AFRNet reduces EER by 6.02, 0.29, and 1.19 points on the left, front, and right subsets, respectively.

\subsection{Synthetic vs.\ Real Data Fine-Tuning}
\label{sec:syn_vs_real}

To investigate the substitution potential of IMPOSE data, we compare fine-tuning with synthetic data versus real contactless data. Real training data uses 500 UWA fingers (outside the test set), comprising $500 \times 4 = 2,000$ plain fingerprints and $500 \times 3 \times 2 = 3,000$ contactless images. DeepPrint and AFRNet, which have fewer spatial alignment constraints, serve as evaluation backbones.

\begin{table}[!t]
\blacktable
  \centering
  \caption{Performance Comparison under Different Fine-Tuning Strategies (\%)}
  \label{tab:ft_compare}
  \resizebox{\linewidth}{!}{
  \begin{threeparttable}
    \begin{tabular}{llcccccc}
      \toprule
      \multirow{2}{*}{\textbf{Method}} & \multirow{2}{*}{\textbf{Strategy}} & \multicolumn{3}{c}{\textbf{UWA}} & \multicolumn{3}{c}{\textbf{PolyU CL2CB}} \\
      \cmidrule{3-5} \cmidrule{6-8}
      & & \textbf{R-1} & \textbf{TAR} & \textbf{EER} & \textbf{R-1} & \textbf{TAR} & \textbf{EER} \\
      \midrule
      \multirow{4}{*}{DeepPrint~\cite{engelsma2021deepprint}} & Baseline & 46.72 & 42.14 & 12.99 & 80.65 & 53.41 & 5.07 \\
      & IMPOSE FT & 64.42 & 62.85 & 10.85 & 93.30 & 81.43 & 3.79 \\
      & UWA FT & 64.93 & 62.99 & 9.10 & 92.61 & 80.23 & 3.22 \\
      & Hybrid FT & 68.37 & 65.96 & \textbf{7.80} & 94.84 & 84.87 & \textbf{2.22} \\
      \midrule
      \multirow{4}{*}{AFRNet~\cite{grosz2023afrnet}} & Baseline & 61.71 & 60.98 & 14.46 & 77.13 & 70.14 & 10.99 \\
      & IMPOSE FT & 67.10 & 66.13 & 11.88 & 85.66 & 77.76 & 8.90 \\
      & UWA FT & 64.29 & 63.39 & 10.78 & 78.82 & 70.47 & 6.39 \\
      & Hybrid FT & 70.18 & 70.88 & \textbf{9.44} & 87.20 & 80.14 & \textbf{4.80} \\
      \bottomrule
    \end{tabular}
    \begin{tablenotes}[flushleft]
\setlength{\labelsep}{0pt}
      \raggedright
      \item[] TAR = TAR@FAR=0.1\%. FT = Fine-Tuned.
    \end{tablenotes}
  \end{threeparttable}}
\end{table}

\begin{figure}[!t]
  \centering
  \subfloat[UWA]{\includegraphics[height=0.4\linewidth]{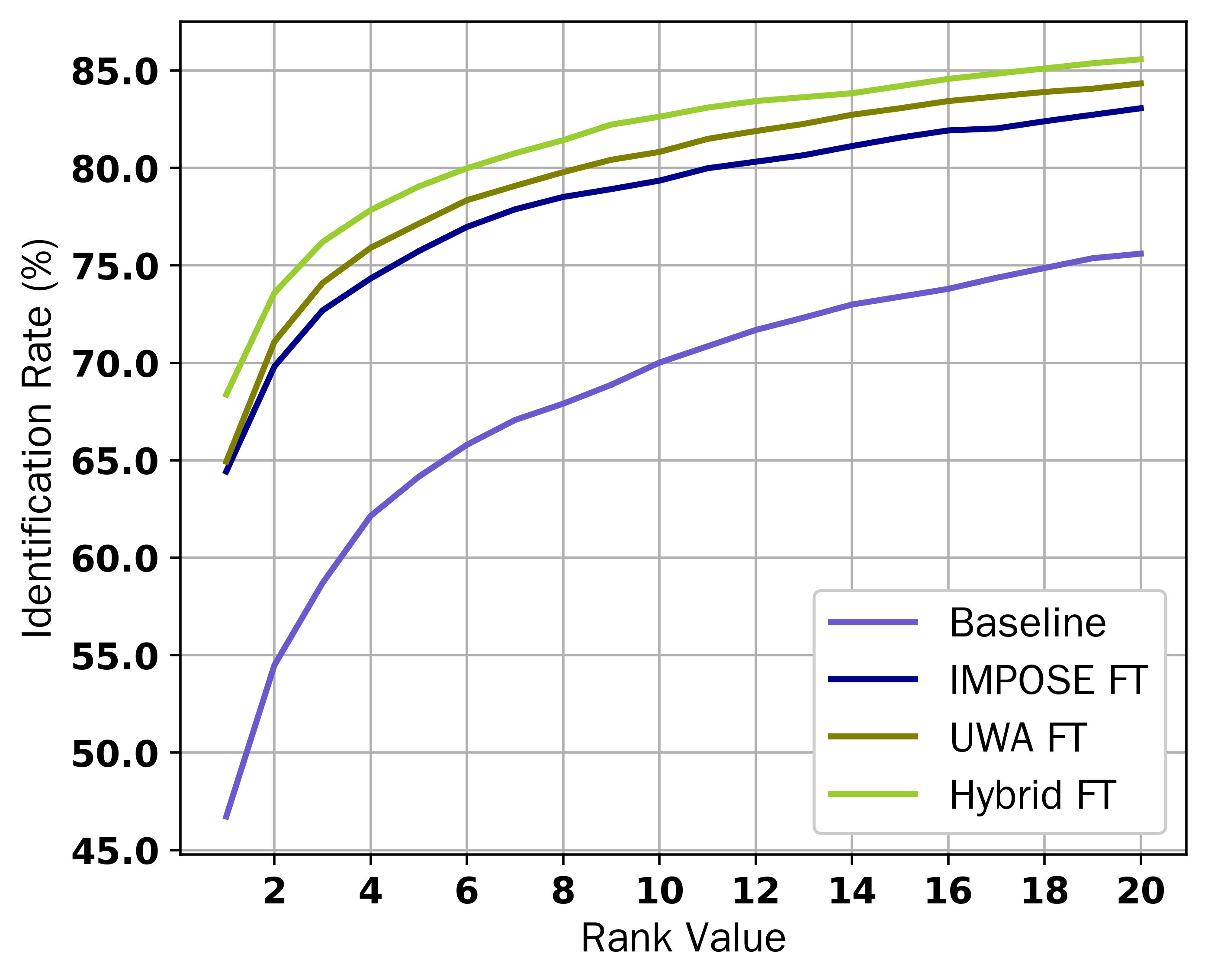}} \hfil
  \subfloat[PolyU CL2CB]{\includegraphics[height=0.4\linewidth]{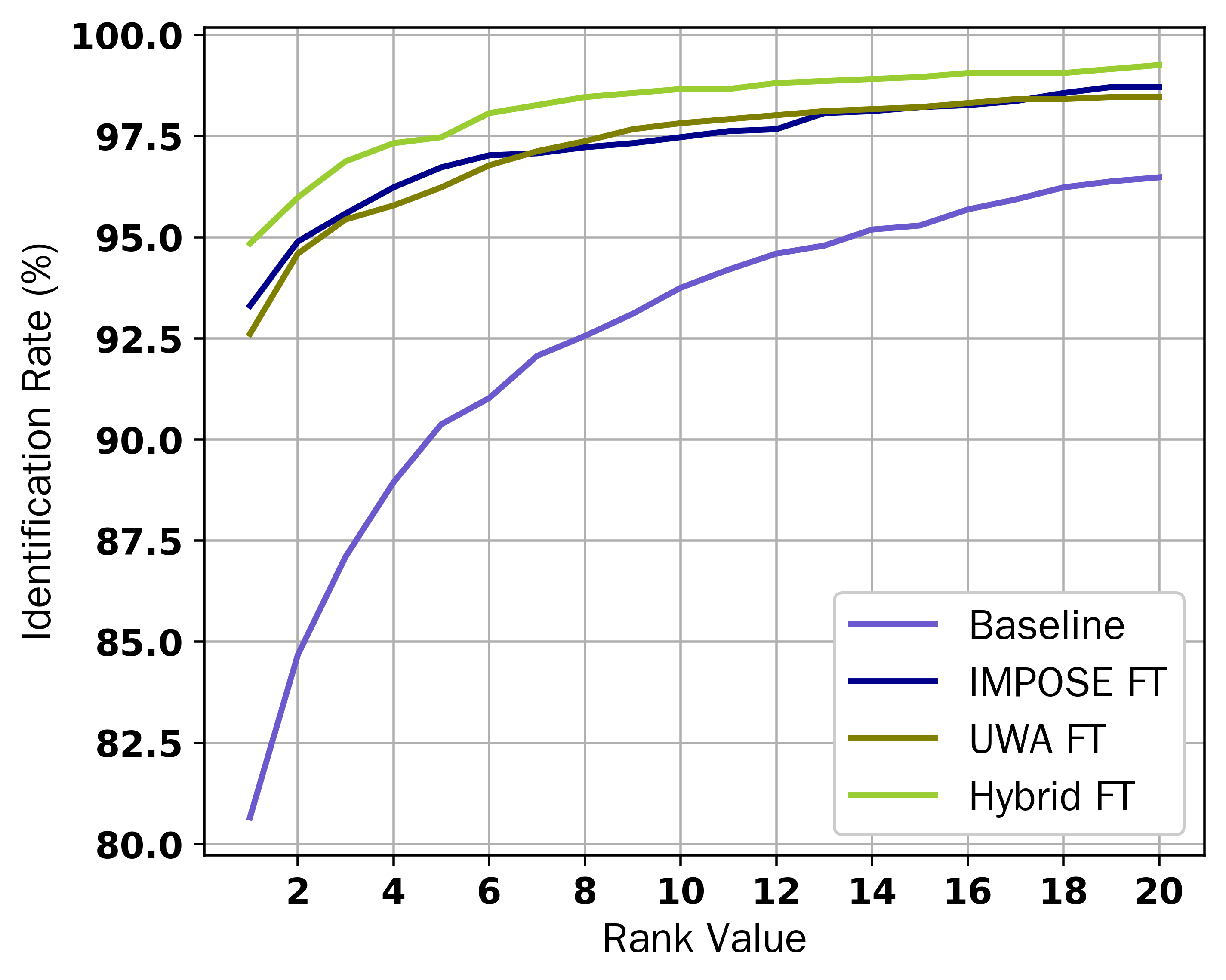}}
  \caption{CMC curves of DeepPrint under different fine-tuning strategies.}
  \label{fig:ft_det_cmc_dp}
\end{figure}

\begin{figure}[!t]
  \centering
  \subfloat[UWA]{\includegraphics[height=0.4\linewidth]{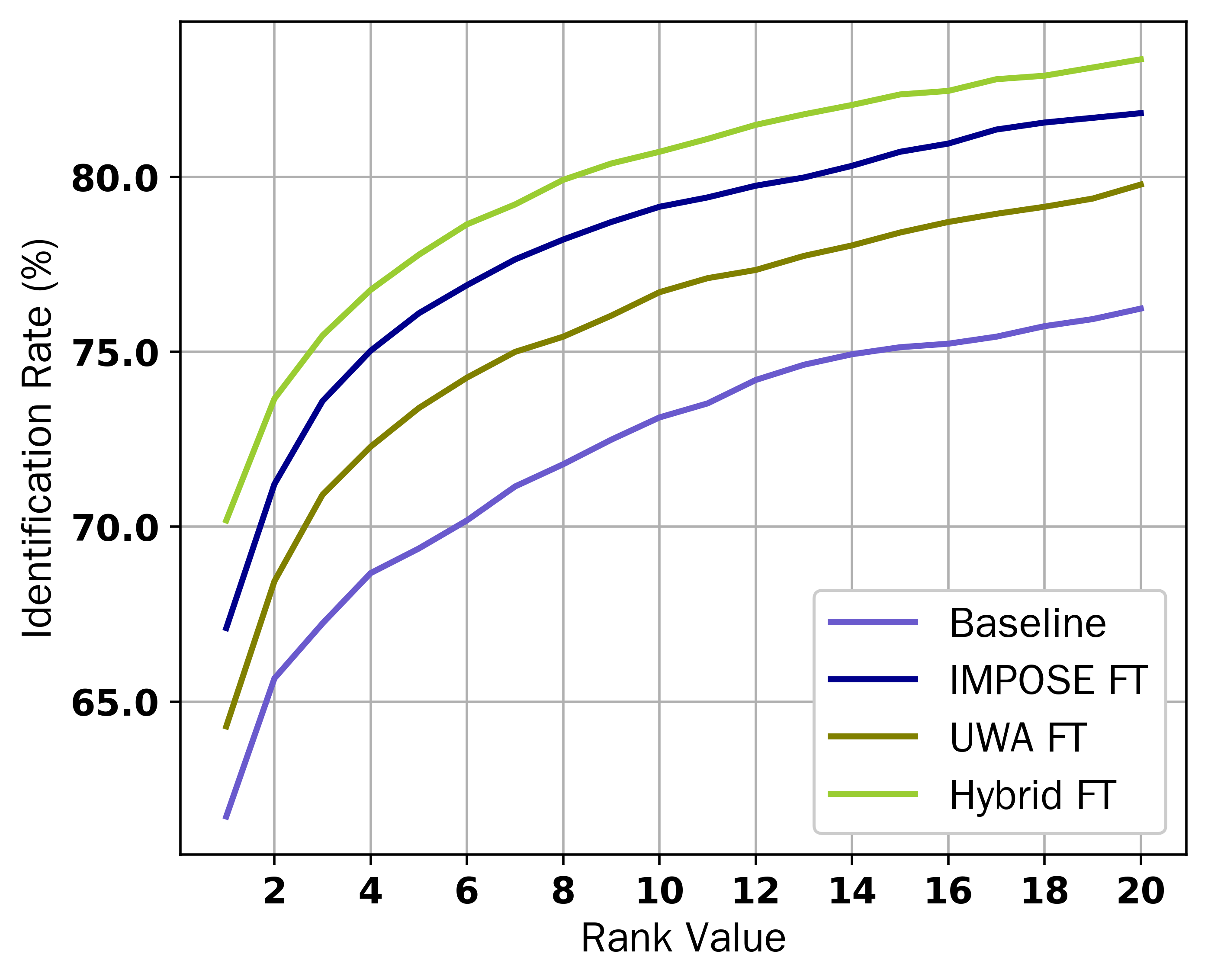}} \hfil
  \subfloat[PolyU CL2CB]{\includegraphics[height=0.4\linewidth]{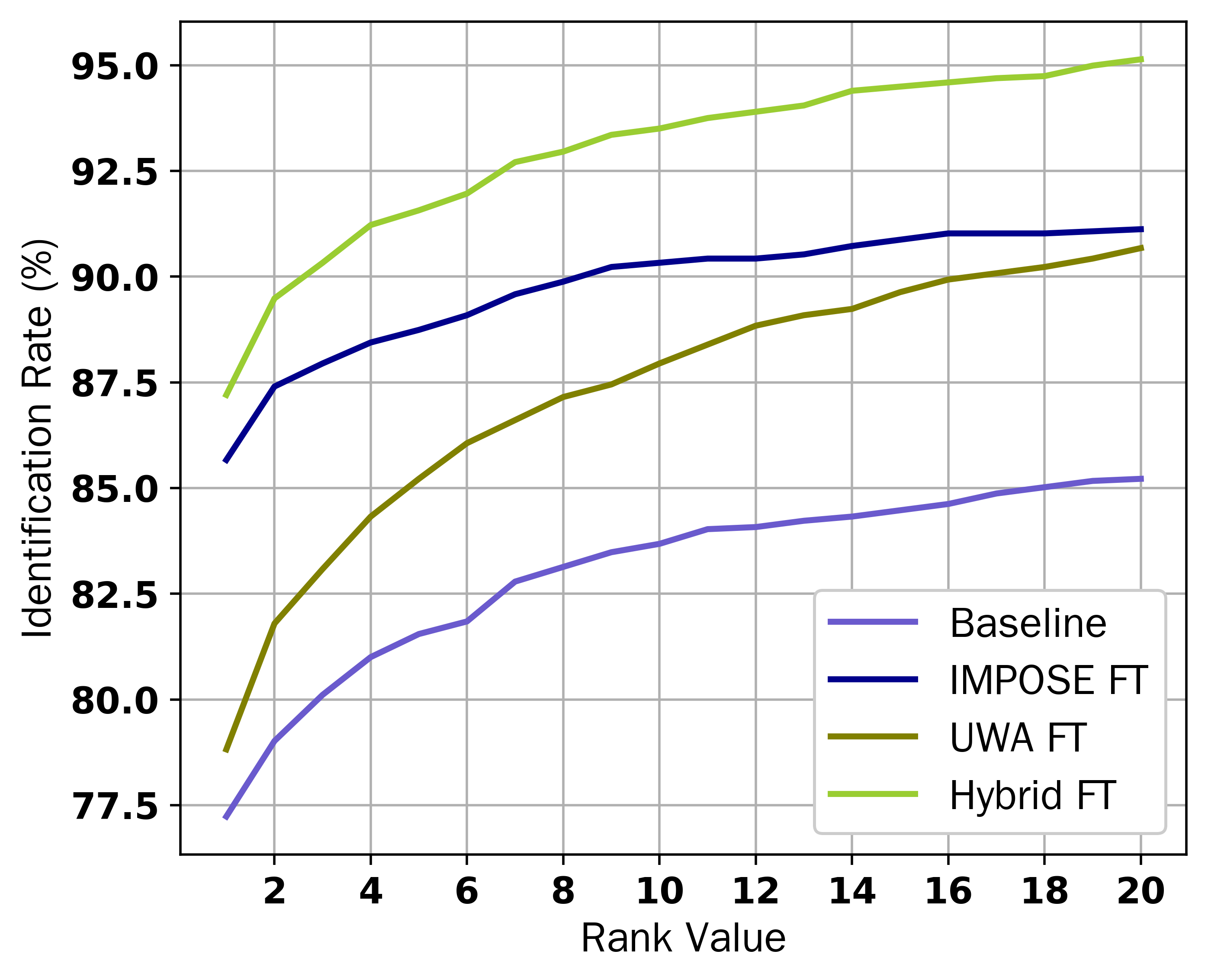}}
  \caption{CMC curves of AFRNet under different fine-tuning strategies.}
  \label{fig:ft_det_cmc_afrnet}
\end{figure}

Table~\ref{tab:ft_compare} shows that fine-tuning solely with IMPOSE data (IMPOSE FT) substantially improves both backbones over their baselines on both databases, including strong cross-dataset gains on PolyU CL2CB. Combining synthetic and real data (Hybrid FT) yields the best overall performance and lowest EERs for both backbones (Figs.~\ref{fig:ft_det_cmc_dp} and~\ref{fig:ft_det_cmc_afrnet}), confirming that IMPOSE and real contactless data are complementary.

\subsection{Comparison at a Nominally Matched Identity Count}
\label{sec:matched500}
\begin{revision}
Table~\ref{tab:matched500} compares FDD fine-tuned with 500 synthetic identities and with 498 real UWA identities (the complete-pair subset of the 500-finger UWA training split) against their common initialization. Both runs start from the same pretrained FDD (identical checkpoint hash), keep the pose estimator fixed, and train the descriptor with two newly initialized classification heads for 60 updates at an effective batch of 48 pairs; the synthetic and real runs use 4,403 and 2,988 contactless images, respectively (Supplementary Table~\suppref{tab:supp_historical_budget}). All three models are evaluated under the protocol of Table~\ref{tab:stage3_ablation}, giving a PolyU EER of 3.19\% for the initial FDD (3.20\% in Table~\ref{tab:comparison}).

\begin{table}[!t]
\blacktable
\revisionfloatcolor
\centering\small\setlength{\tabcolsep}{3pt}
\caption{\rev{FDD Fine-Tuned with Nominally 500 Identities and Its Initialization (\%)}}
\label{tab:matched500}
\begin{threeparttable}
\begin{tabular}{llrrrr}
\toprule
Test&Training&R-1&TAR$_{.1}$&TAR$_{.01}$&EER\\
\midrule
UWA&Initial FDD&73.33&73.39&63.49&12.01\\
&IMPOSE&66.83&66.33&56.33&15.03\\
&Real UWA&68.54&68.34&59.91&14.93\\
\midrule
PolyU&Initial FDD&91.82&88.17&79.82&3.19\\
&IMPOSE&91.07&84.76&76.63&4.02\\
&Real UWA&84.67&76.57&66.06&7.66\\
\bottomrule
\end{tabular}
\begin{tablenotes}[flushleft]
\setlength{\labelsep}{0pt}
\footnotesize\raggedright
\item[] R-1 = Rank-1. TAR$_{.1}$ and TAR$_{.01}$ denote TAR at FAR=0.1\% and 0.01\%, respectively.
\end{tablenotes}
\end{threeparttable}
\end{table}

Both 60-update models fall below the initial FDD on every metric, which explains why their EERs exceed the FDD baseline in Table~\ref{tab:comparison}. We attribute this drop to the fixed pose estimator and the short 60-update schedule: early gradients from the newly initialized heads perturb the pretrained descriptor before head convergence~\cite{kumar2022finetuning}. Notably, IMPOSE data preserves cross-dataset generalization on PolyU much better than real UWA data (Rank-1 $-0.75$ and EER $+0.83$ points vs.\ $-7.15$ and $+4.47$ points, giving 6.40-point higher Rank-1 and 3.64-point lower EER), as its wider pose coverage mitigates overfitting to UWA capture conditions, whereas real UWA data degrades slightly less on UWA.
\end{revision}

\section{Discussion and Limitations}

{Fine-tuning with IMPOSE data improves FDD~\cite{pan2024FDD}, DeepPrint~\cite{engelsma2021deepprint}, and AFRNet~\cite{grosz2023afrnet} in the main experiments, and 3D-projected views outperform 2D TPS augmentation under an equal budget. As shown in Sections~\ref{sec:fdd_enhance}--\ref{sec:stage3_ablation} and~\ref{sec:matched500}, realizing the full gain on FDD requires joint pose-estimator and descriptor adaptation over a sufficient schedule. Several limitations remain.}

\textbf{Pose Coverage.} The quantitative evaluation focuses on finger roll, which drives the nonlinear ridge deformation and visibility changes emphasized by the UWA protocol. Although the same 3D representation can be rotated around other axes, recognition gains for pitch and yaw are not evaluated here. Unconstrained mobile capture also introduces camera-distance changes, illumination variation, out-of-plane bending, and external occlusion. Future simulation should jointly vary these factors to assess robustness under realistic free-hand acquisition.

\textbf{3D Model Diversity.} The geometric library contains 285 measured finger point clouds that provide subject-specific surface shape but are treated as rigid templates during projection. They do not span population variation in finger thickness, curvature, skin elasticity, creases, or pressure-dependent deformation. Broader measured collections and deformable or parametric surface models could increase geometric diversity while preserving the texture-to-surface correspondence used by IMPOSE.

{
\textbf{Offline deployment.} IMPOSE operates offline to generate training data; the diffusion models and 3D projection stage are not executed during descriptor-based authentication. {Consequently, the synthesis timings in Supplementary Table~\suppref{tab:supp_timing} do not add to authentication latency, which is determined by the underlying matcher alone.} Faster projection or generative-model distillation would mainly accelerate dataset construction and iterative training rather than runtime recognition.

\textbf{Generation failures.} Figure~\ref{fig:failure_summary} shows two rejection mechanisms: NFIQ rejection and insufficient foreground after passing NFIQ. {Supplementary Figs.~\suppref{fig:supp_failure_nfiq} and~\suppref{fig:supp_failure_mask} provide additional examples;} these cases are illustrative rather than estimates of failure prevalence.
\begin{figure}[!t]
\centering
\subfloat[NFIQ-rejected rolled samples.\label{fig:failure_summary_a}]{\includegraphics[width=.90\linewidth]{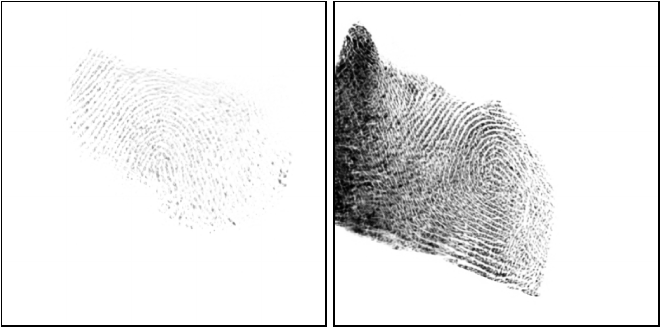}}\\
\subfloat[Low-foreground rejection after NFIQ.\label{fig:failure_summary_b}]{\includegraphics[width=.90\linewidth]{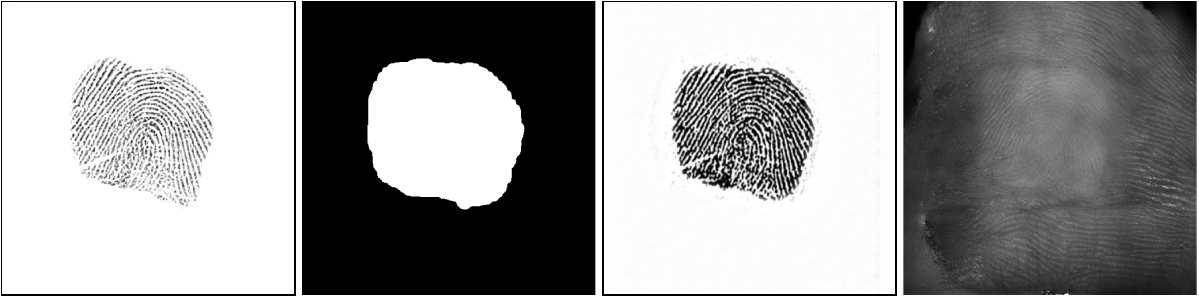}}
\caption{Representative rejected samples in the IMPOSE generation pipeline.}\label{fig:failure_summary}
\end{figure}
}

\begin{revision}
\textbf{Large roll angles.} Projection preserves the ridge pattern but not its visibility. Beyond $\pm30^\circ$, part of the ridge area turns away from the camera and the rest is foreshortened, lowering Rank-1 by 2.4--3.3 points relative to the frontal view (Section~\ref{sec:identity_quant}).
\end{revision}

\textbf{From Simulation to Generation.} A natural next direction is moving from explicit geometric rendering toward end-to-end 3D-aware generation. Integrating physical surface priors directly with diffusion models could synthesize contactless images with more realistic illumination and non-rigid skin deformation under complex capture conditions.

\section{Conclusion}

\begingroup
This paper presented IMPOSE, which combines rolled-identity synthesis, Sauvola-guided contactless translation, and physics-based 3D multi-pose projection. The translated images match their rolled sources with 100\% Rank-1 for both DMD and VeriFinger, and after projection Rank-1 remains 98.57\% and 98.64\%, with most errors at roll angles beyond $\pm30^\circ$. Fine-tuning with IMPOSE data improves FDD, DeepPrint, and AFRNet on UWA and PolyU CL2CB, and combining synthetic and real data gives the lowest EERs for DeepPrint and AFRNet. Under an equal training budget, projected views outperform 2D TPS augmentation in Rank-1 and EER. Because appearance synthesis and pose simulation are decoupled, improved contactless rendering or richer 3D geometry can be incorporated without changing the descriptor. Future work will extend the simulation to pitch, yaw, and non-rigid deformation and enlarge the 3D finger library.
\endgroup

\bibliographystyle{styles/IEEEtran}
\bibliography{bib/CL_generation,bib/descriptor,bib/others,bib/preprocessing,bib/revision_references}

\input{sections/supplementary.tex}

\end{document}

%% file: sections/supplementary.tex

\clearpage
\raggedbottom
\setcounter{section}{7}
\setcounter{table}{0}
\setcounter{figure}{0}
\renewcommand{\thetable}{S\arabic{table}}
\renewcommand{\thefigure}{S\arabic{figure}}
\makeatletter
\setlength{\@fptop}{0pt}
\setlength{\@fpsep}{12pt}
\makeatother

\twocolumn[{\centering{\Large\bfseries Supplementary Material}\par\vspace{0.6\baselineskip}}]

\begin{revision}
\section{Detailed Implementation and Reproducibility Notes}
This supplement provides the autoencoder objectives and architectures, angle-stratified identity evaluation of the projected images, protocols of the controlled and 500-identity fine-tuning experiments, and computational cost.
\end{revision}

\input{sections/supp_method.tex}

\begin{revision}
\subsection{Parameter Counts and Timing}
Table~\ref{tab:supp_params} lists the parameter count and training time of each component.

\begin{table}[!t]
\blacktable
\revisionfloatcolor
\centering\footnotesize\setlength{\tabcolsep}{2.5pt}
\caption{\rev{Component Parameter Counts and Training Durations}}
\label{tab:supp_params}
\begin{tabular}{lccc}
\toprule
Module & Total (M) & Optimized (M) & Duration (h)\\
\midrule
Rolled VQ-VAE &13.856&13.856&4.4\\
Rolled LDM U-Net &16.334&16.334&12.9\\
Sauvola-map AE &13.843&13.843&10.0\\
Contactless VQ-VAE &13.856&13.856&2.1\\
Contactless LDM U-Net &16.334&16.334&4.2\\
ControlNet side branch &7.040&7.040&5.2\\
\bottomrule
\end{tabular}
\end{table}

\begin{table}[!t]
\blacktable
\revisionfloatcolor
\centering\footnotesize\setlength{\tabcolsep}{3pt}
\caption{\rev{Offline Generation Timings}}
\label{tab:supp_timing}
\begin{threeparttable}
\begin{tabular}{lrr}
\toprule
Benchmark & Training (h) & Output rate (images/s)\\
\midrule
Rolled generation &17.3&2.26\\
Translation &21.5&1.36\\
CPU projection &N/A&0.087\\
\bottomrule
\end{tabular}
\begin{tablenotes}[flushleft]
\setlength{\labelsep}{0pt}
\scriptsize\raggedright
\item[] N/A = not applicable.
\end{tablenotes}
\end{threeparttable}
\end{table}

Training time in Table~\ref{tab:supp_timing} sums the corresponding submodules in Table~\ref{tab:supp_params}. Generation latency is measured on an RTX~4090 GPU for $512\times512$ outputs using 50 DDIM steps and a batch size of 1, excluding model loading. Cross-modal translation takes $0.00835\pm0.00008$\,s of CPU Sauvola preprocessing and $0.7246\pm0.0018$\,s of 50-step DDIM sampling per image (1.36 images/s). Projecting one contactless texture into nine views on the CPU takes $103.362\pm0.086$\,s (0.087 views/s).

Training durations are obtained from the training logs (with the 120-epoch Sauvola-AE duration estimated from its per-epoch runtime). All modules are trained on one RTX~4090 except the Sauvola AE, which uses two GPUs with a global batch of 16.
\end{revision}

\input{sections/supp_identity.tex}

\begin{revision}
\section{Controlled and 500-Identity Fine-Tuning Protocols}
\end{revision}
\input{sections/supp_training.tex}

\section{\rev{Additional Qualitative Diagnostic Results}}

\subsection{Sauvola Parameter Sensitivity}
\begin{figure}[!t]
\centering
\subfloat[Normal]{\includegraphics[width=.70\linewidth]{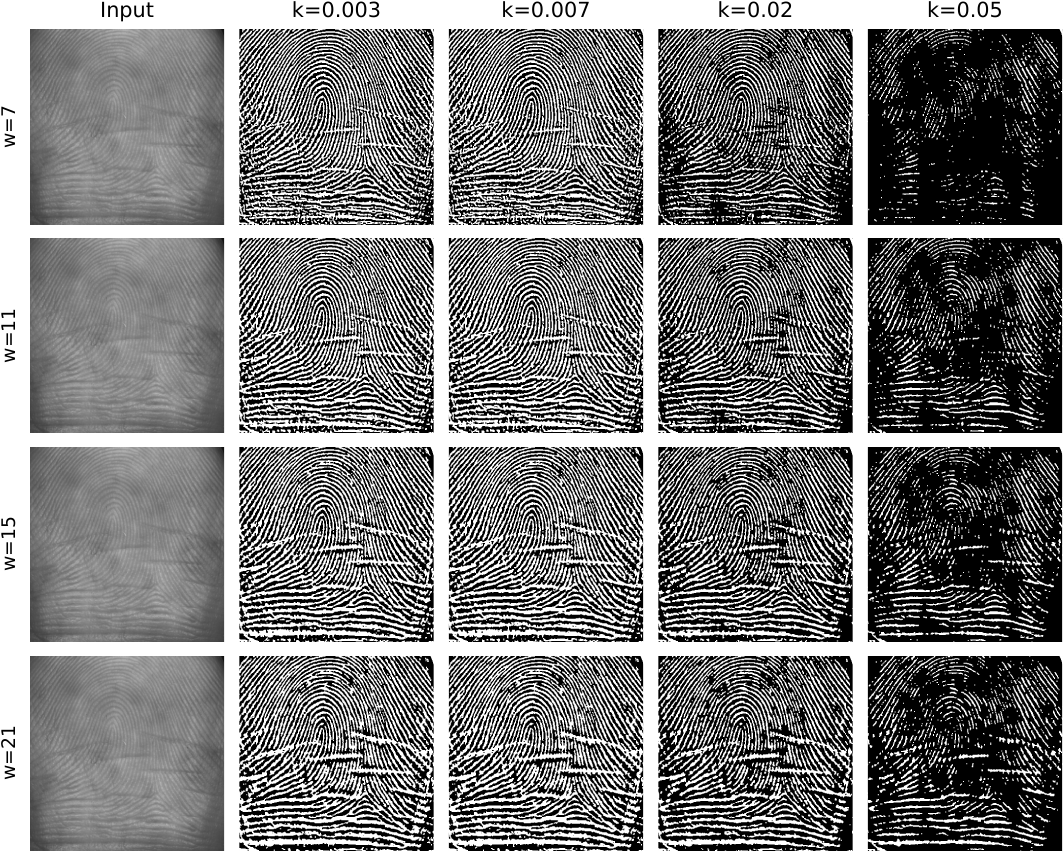}}\\[1mm]
\subfloat[Low contrast]{\includegraphics[width=.70\linewidth]{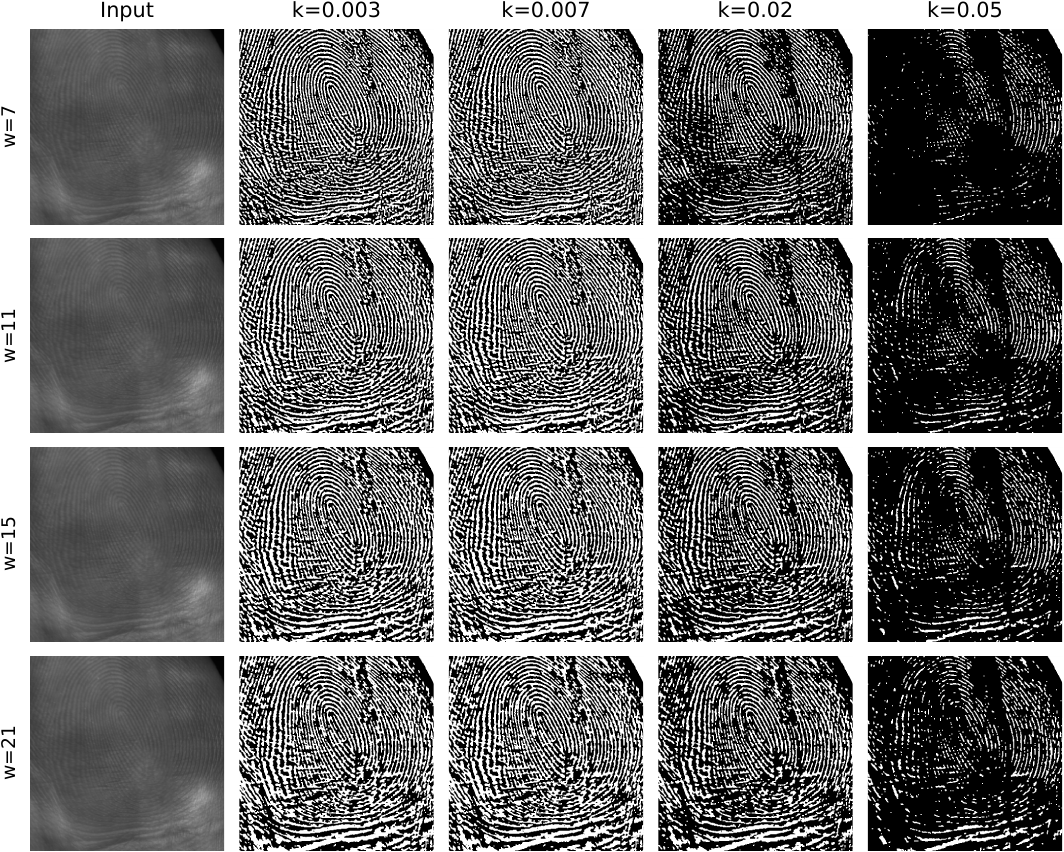}}\\[1mm]
\subfloat[Local noise]{\includegraphics[width=.70\linewidth]{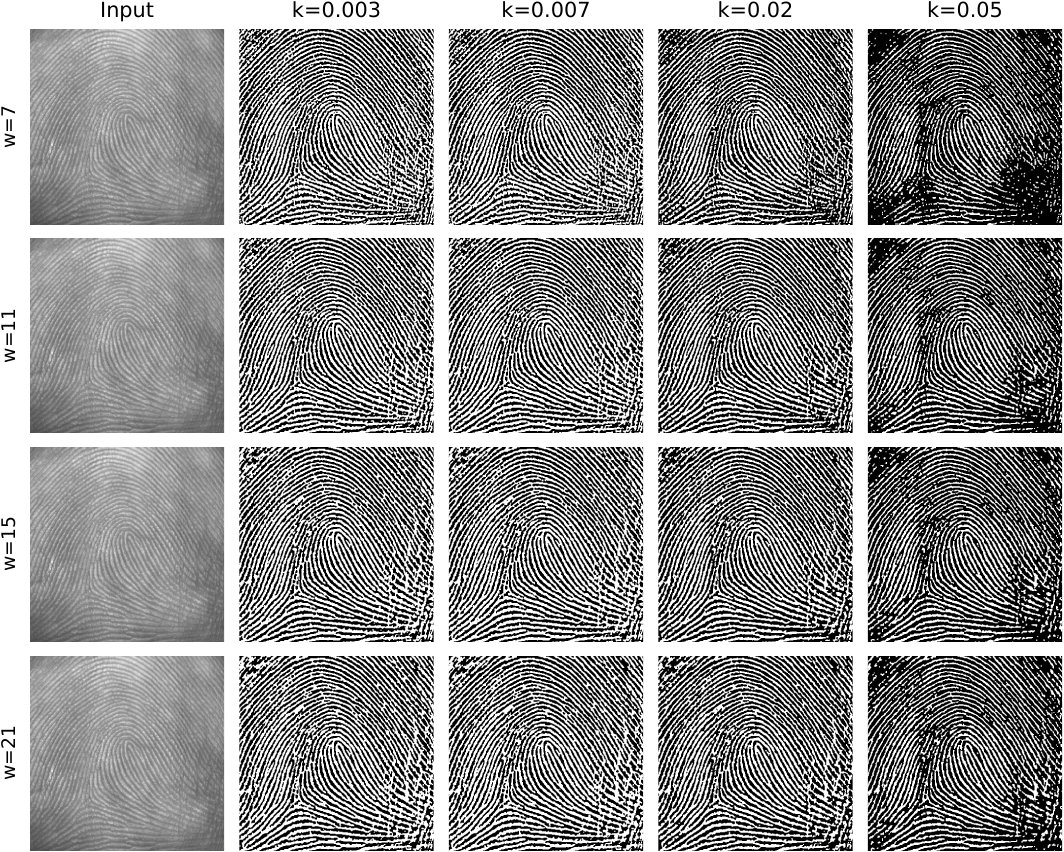}}
\caption{Complete Sauvola parameter-sensitivity grids for three representative contactless samples; rows vary window size $w$ and columns vary correction factor $k$.}
\label{fig:supp_sauvola}
\end{figure}

The main paper summarizes the sensitivity of the Sauvola conditioning parameters. Figure~\ref{fig:supp_sauvola} provides the complete qualitative grids for the three representative samples. The same settings are used in every panel: $w\in\{7,11,15,21\}$, $k\in\{0.003,0.007,0.02,0.05\}$, and $R=128$. Small $k$ values preserve similar ridge topology across the tested window sizes, while larger $k$ values progressively darken local regions and merge neighboring ridges.

\begin{figure}[!t]
\centering
\subfloat[Weak ridges (ID 00016).]{\includegraphics[width=.31\linewidth]{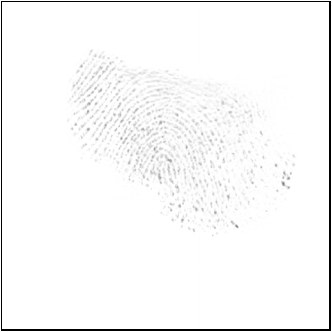}}\hfill
\subfloat[Incomplete coverage (ID 00126).]{\includegraphics[width=.31\linewidth]{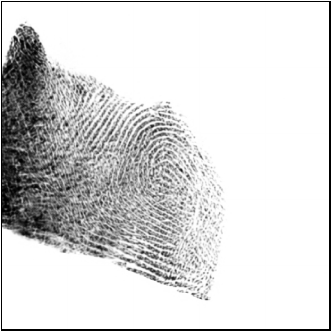}}\hfill
\subfloat[Blurred texture (ID 00162).]{\includegraphics[width=.31\linewidth]{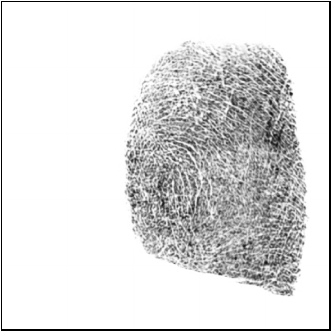}}
\caption{Rolled-generation failures rejected by NFIQ; all three examples have an NFIQ 2.0 score of 0.}
\label{fig:supp_failure_nfiq}
\end{figure}

\begin{figure}[!t]
\centering
\subfloat[ID 09030: NFIQ 58, mask ratio 0.22.]{\includegraphics[width=.95\linewidth]{figures/revision/highres/failure_mask_09030_bordered.pdf}}\\[1.0mm]
\subfloat[ID 04030: NFIQ 57, mask ratio 0.22.]{\includegraphics[width=.95\linewidth]{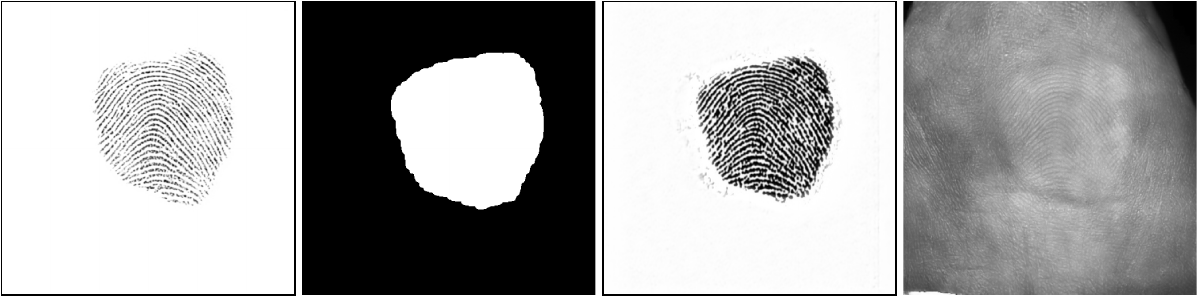}}\\[1.0mm]
\subfloat[ID 09547: NFIQ 62, mask ratio 0.23.]{\includegraphics[width=.95\linewidth]{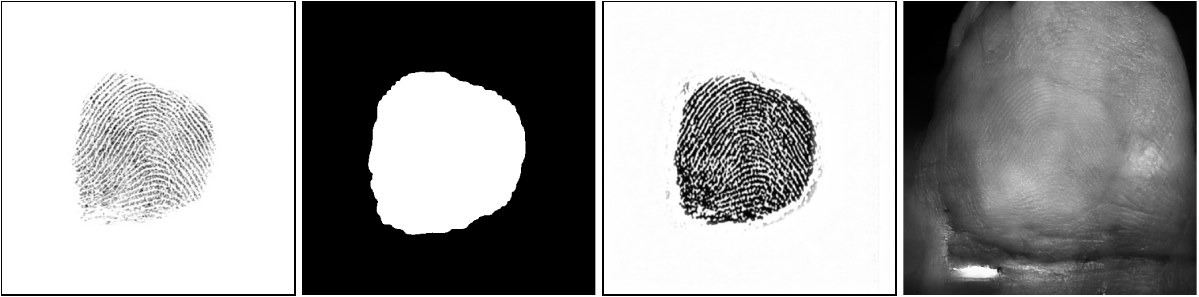}}
\caption{NFIQ-passing identities rejected for low foreground. Columns show the rolled image, foreground mask, Sauvola condition, and Stage 2 output.}
\label{fig:supp_failure_mask}
\end{figure}

\subsection{Additional Generation Failure Examples}
{\clubpenalty=0 Figures~\ref{fig:supp_failure_nfiq} and~\ref{fig:supp_failure_mask} expand the two representative rejection mechanisms shown in the main text. They are selected examples from the corresponding filtering stages and are not used to estimate failure prevalence.\par}

\Needspace{12\baselineskip}
\section{Additional Notes on Simulation Scope}
\leavevmode\rev{The final Stage~3 evaluation is stratified by roll angle. Only roll is varied; pitch, yaw, camera distance, illumination, non-rigid bending, and external occlusion are not evaluated.} The rigid library contains 285 measured finger surfaces but does not cover full anatomical or non-rigid diversity; broader measured collections and deformable or parametric surface models remain future work.

%% file: sections/supp_method.tex
\begin{revision}
\subsection{Detailed Architecture and Sampling Configuration}
\label{sec:supp_architecture}

Table~\ref{tab:supp_arch} specifies the image autoencoders, diffusion backbone, and conditioning path. The rolled and contactless VQ-VAEs use the same block layout and codebook size. The Sauvola autoencoder shares their encoder--decoder layout but uses a continuous bottleneck, without vector quantization. Thus its binary input and reconstruction have size $1\times512\times512$, and its condition latent has size $3\times128\times128$. The rolled generator is unconditional; contactless translation is conditioned on this latent rather than on the binary image directly.
\end{revision}

\input{sections/supp_arch_table.tex}

\begin{revision}
The spatial resolutions in Table~\ref{tab:supp_arch} follow the two image-encoder downsampling operations. The diffusion models use AdamW and a linear noise schedule with 1,000 training steps; generation uses 50 DDIM steps. Both Eq.~(\mainref{eq:ddpm_loss}) and the released code use the $\ell_1$ noise-prediction loss.

\subsection{VQ-VAE Pretraining Objective}
\label{sec:supp_vq_loss}

For both image VQ-VAEs, the training objective combines pixel reconstruction, perceptual reconstruction, and vector quantization:
\begin{equation}
  \mathcal{L}_{\mathrm{AE}}
  = \mathcal{L}_{\ell_1}
  + \mathcal{L}_{\mathrm{LPIPS}}
  + \mathcal{L}_{\mathrm{VQ}}.
  \label{eq:supp_vq_total}
\end{equation}
All three outer weights are $1.0$. The commitment coefficient within $\mathcal{L}_{\mathrm{VQ}}$ is $\beta=0.25$, separate from its unit outer weight. Table~\ref{tab:supp_vq_loss} records both the active losses and the disabled options. In particular, no adversarial or fingerprint-identity term contributes to VQ-VAE pretraining.

\begin{table}[!t]
\blacktable
\revisionfloatcolor
\centering
\footnotesize
\setlength{\tabcolsep}{3pt}
\caption{\rev{VQ-VAE Pretraining Losses and Switches}}
\label{tab:supp_vq_loss}
\begin{tabular}{>{\raggedright\arraybackslash}p{.31\columnwidth}>{\raggedright\arraybackslash}p{.25\columnwidth}>{\raggedright\arraybackslash}p{.34\columnwidth}}
\toprule
Term / option & Weight / setting & Role in training\\
\midrule
Pixel reconstruction & $\ell_1$; $1.0$ & Active\\
Perceptual reconstruction & LPIPS; $1.0$ & Active\\
Vector quantization & $1.0$ & Active\\
Commitment term & $\beta=0.25$ & Inside $\mathcal{L}_{\mathrm{VQ}}$\\
Adversarial loss & \makecell[l]{\texttt{use\_gan}\\\texttt{=False}} & Disabled\\
Fingerprint identity loss & \makecell[l]{\texttt{finger\_loss}\\\texttt{=False}} & Disabled\\
Discriminator settings & Weight $0.5$; start $4001$ & Inactive because GAN is disabled\\
\bottomrule
\end{tabular}
\end{table}

\subsection{Sauvola Autoencoder and ControlNet Connection}
\label{sec:supp_sauvola_control}

The Sauvola autoencoder is first trained to reconstruct a binary ridge map $B$ using the mean squared error between $B$ and $D_{\mathrm e}(E_{\mathrm e}(B))$. Its encoder has channel widths $64$, $128$, and $256$, with two residual blocks per level and two factor-of-two downsampling operations. A residual--attention--residual bottleneck produces the continuous representation. The decoder reverses the channel sequence, uses three residual blocks per level and two factor-of-two upsampling operations, and reconstructs a one-channel binary image. Table~\ref{tab:supp_arch} gives the corresponding resolutions.

During cross-modal translation, the conditioning path is
\begin{equation}
  B \xrightarrow{\,E_{\mathrm e}\ \mathrm{(frozen)}\,}
  z_B \in \mathbb{R}^{3\times128\times128}.
  \label{eq:supp_sauvola_latent}
\end{equation}
ControlNet processes $z_B$ as a spatial condition together with the noisy contactless latent and diffusion timestep. Its multiscale control features pass through zero-initialized convolutional connections into the pretrained denoising U-Net. The zero initialization prevents the control branch from perturbing the pretrained backbone at initialization. The binary reconstruction decoder $D_{\mathrm e}$ is used only for autoencoder pretraining and does not participate in translation.

The encoder $E_{\mathrm e}$ and pretrained contactless denoising U-Net are frozen during ControlNet training; only the ControlNet side branch is optimized. Training uses $B_{\mathrm{cl}}=\operatorname{Sauvola}(I_{\mathrm{cl}})$ from the target contactless image. At generation, the same encoder and conditioning path receive $B_r=\operatorname{Sauvola}(I_r)$ from a generated rolled fingerprint. These stages therefore share the map-to-latent connection while using different source modalities for the map. There is no additional identity loss in this translation stage.
\end{revision}

%% file: sections/supp_arch_table.tex
\begin{table*}[!t]
\blacktable
\centering\footnotesize
\setlength{\tabcolsep}{4pt}
\renewcommand{\arraystretch}{1.05}
\caption{\rev{Block-Level Architecture and Conditioning Path}}
\label{tab:supp_arch}
\begin{tabular}{>{\raggedright\arraybackslash}p{.23\textwidth}>{\raggedright\arraybackslash}p{.18\textwidth}>{\raggedright\arraybackslash}p{.51\textwidth}}
\toprule
Block & Output / channels & Structure and operation\\
\midrule
\multicolumn{3}{l}{\textbf{Shared rolled/contactless VQ-VAE}}\\
Input convolution & $512^2$, 64 & $3\!\times\!3$, $1\!\rightarrow\!64$\\
Encoder level 0 & $512^2$, 64 & 2 ResBlocks; conv. downsample\\
Encoder level 1 & $256^2$, 128 & 2 ResBlocks; conv. downsample\\
Encoder level 2 & $128^2$, 256 & 2 ResBlocks\\
Encoder middle & $128^2$, 256 & ResBlock--attention--ResBlock\\
Latent / quantizer & $128^2$, 3 & $1\!\times\!1$ projection; 4,096 codes of width 3\\
Decoder middle & $128^2$, 256 & ResBlock--attention--ResBlock\\
Decoder levels 2/1/0 & $128^2\!\to\!512^2$ & 3 ResBlocks/level; channels 256/128/64; inter-level upsampling\\
Output convolution & $512^2$, 1 & norm., activation, $3\!\times\!3$ conv.\\
\addlinespace[2pt]
\multicolumn{3}{l}{\textbf{Shared latent-diffusion U-Net}}\\
Input / time embed. & $128^2$, 64 & $3\!\rightarrow\!64$ conv.; MLP $64\!\rightarrow\!256\!\rightarrow\!256$\\
Down levels 0/1/2 & $128^2/64^2/32^2$ & 2 ResBlocks/level; channels 64/128/256; ResBlock downsampling\\
Middle & $32^2$, 256 & ResBlock--8-head attention--ResBlock\\
Up levels 2/1/0 & $32^2/64^2/128^2$ & 3 skip-connected ResBlocks/level; channels 256/128/64\\
Output & $128^2$, 3 & norm., SiLU, $3\!\times\!3$ conv.\\
Diffusion / sampling & --- & 1,000 linear-noise steps; $\ell_1$ prediction; DDIM-50 ($\eta=0$ for rolled)\\
Attention setting & --- & Configured list \texttt{[32,16,8]}, 8 heads; middle attention instantiated\\
Training status & --- & U-Net optimized; VQ-VAE frozen\\
\addlinespace[2pt]
\multicolumn{3}{l}{\textbf{Sauvola-map continuous autoencoder}}\\
Input & $512^2$, 1 & Sauvola binary ridge map $B$\\
Encoder levels 0/1/2 & $512^2/256^2/128^2$ & 2 ResBlocks/level; channels 64/128/256; two factor-of-two downsampling operations\\
Encoder middle & $128^2$, 256 & ResBlock--attention--ResBlock\\
Continuous latent & $128^2$, 3 & $z_B=E_{\mathrm e}(B)$; no vector quantization\\
Decoder levels 2/1/0 & $128^2\!\to\!512^2$ & 3 ResBlocks/level; channels 256/128/64; two factor-of-two upsampling operations\\
Reconstruction / objective & $512^2$, 1 & $D_{\mathrm e}(z_B)$; mean squared reconstruction error\\
Translation-stage status & Encoder frozen & $D_{\mathrm e}$ does not participate in translation\\
\addlinespace[2pt]
\multicolumn{3}{l}{\textbf{ControlNet branch}}\\
Spatial condition & $128^2$, 3 & Continuous Sauvola-map latent $z_B$ processed by the control branch\\
Controlled encoder & $128^2/64^2/32^2$ & mirrors U-Net down path; 2 ResBlocks/level\\
Control outputs & 9 encoder + 1 middle & zero $1\!\times\!1$ convs.; middle-only control disabled\\
Training status & --- & ControlNet optimized; Sauvola encoder and denoising backbone frozen\\
\bottomrule
\end{tabular}
\end{table*}

%% file: sections/supp_identity.tex
\begin{revision}
\section{Identity Preservation Before and After Projection}
\label{sec:supp_identity}

\subsection{Stage 2 Translation Diagnostics}
The translation-only evaluation uses one frontal Stage 2 contactless query per retained identity and a gallery of 1,712 generated rolled fingerprints. Both matchers evaluate all 1,712 genuine and 2,929,232 impostor pairs, without impostor subsampling. Table~\ref{tab:supp_stage2_identity} gives the detailed results that complement the final-projection evaluation in the main text. Matching after Stage~3 projection is evaluated in Sections~\ref{sec:supp_projection_set}--\ref{sec:supp_large_angle}.

\begin{table}[!t]
\blacktable
\revisionfloatcolor
\centering\footnotesize
\setlength{\tabcolsep}{3pt}
\caption{\rev{Stage 2 Translation-Only Identity Preservation (\%)}}
\label{tab:supp_stage2_identity}
\begin{threeparttable}
\begin{tabular}{lrrrr}
\toprule
Matcher & R-1/5/10 & TAR & Actual FAR & EER\\
\midrule
DMD &100/100/100&100&0.0995483&0.0000341\\
VeriFinger &100/100/100&100&0.0778361&0\\
\bottomrule
\end{tabular}
\begin{tablenotes}[flushleft]
\setlength{\labelsep}{0pt}
\scriptsize\raggedright
\item[] TAR at an empirical FAR $\le$ 0.1\%.
\end{tablenotes}
\end{threeparttable}
\end{table}

DMD genuine scores have mean $0.088480\pm0.038286$ and range $0.016650$--$0.244048$; impostor scores have mean $0.006465\pm0.001358$ and range $0.001988$--$0.017723$. Only two impostor scores reach or exceed the minimum genuine score, and the correct identity ranks first for every query. VeriFinger genuine scores have mean $249.032\pm64.076$ and range $60$--$442$, compared with impostor scores of $6.034\pm5.589$ and range $0$--$57$. Figure~\ref{fig:supp_stage2_identity} shows the corresponding DET curves and score separation. The evaluation covers the 1,712 identities that pass quality filtering.

\begin{figure}[!t]
\revisionfloatcolor
\centering
\subfloat[Low-error DET curves.]{\includegraphics[width=.90\linewidth]{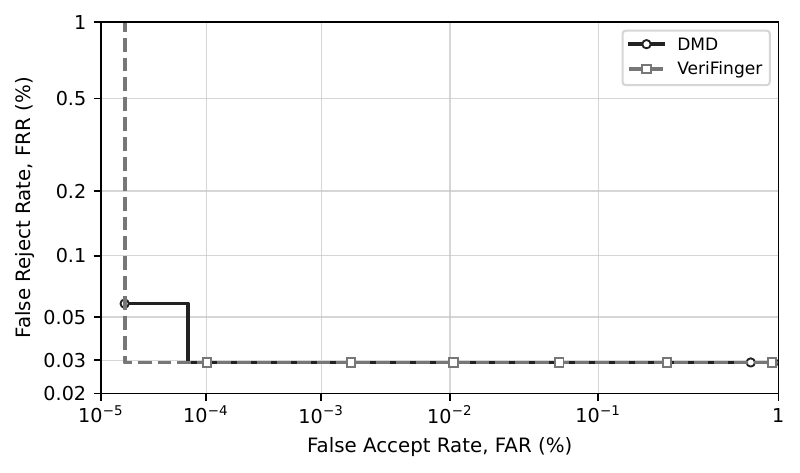}}\\
\subfloat[Genuine/impostor score separation.]{\includegraphics[width=.90\linewidth]{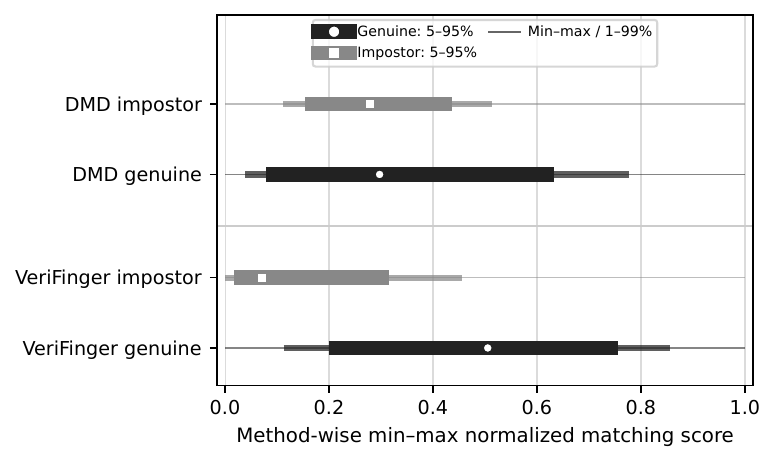}}
\caption{\rev{Stage 2 translation-only evaluation. Scores in (b) are normalized separately for each matcher. Zero-error points in (a) are displayed at half-count floors.}}
\label{fig:supp_stage2_identity}
\end{figure}

\subsection{Stage 3 Evaluation Set}
\label{sec:supp_projection_set}
The final-projection evaluation uses all 15,136 final Stage 3 images as queries and the same 1,712 rolled fingerprints as gallery references. Each query is matched against the entire gallery. Minutiae are extracted independently from every image; genuine pairs share an identity and impostor pairs have different identities.

The schedule requests $1{,}712\times9=15{,}408$ views. Because non-frontal angles are drawn with replacement and files are named by identity and angle, a repeated draw overwrites an existing file. The resulting 15,136 images (Table~\ref{tab:supp_projection_counts}) include a $0^\circ$ view for every identity; all 272 missing images are repeated draws (one repeat for each of 252 identities and two for each of 10), and no projection failed.

\begin{table}[!t]
\blacktable
\revisionfloatcolor
\centering\footnotesize
\setlength{\tabcolsep}{5pt}
\caption{\rev{Stage 3 Projected Views per Identity}}
\label{tab:supp_projection_counts}
\begin{tabular}{rrr}
\toprule
Views per identity & Identities & Images\\
\midrule
9 &1,450&13,050\\
8 &252&2,016\\
7 &10&70\\
\midrule
Total &1,712&15,136\\
\bottomrule
\end{tabular}
\end{table}

DMD cannot score one query (Section~\ref{sec:supp_large_angle}); VeriFinger scores all 15,136 genuine and 25,897,696 impostor comparisons. The DMD group from $-59^\circ$ to $-31^\circ$ has 3,377 valid genuine scores for 3,378 attempted queries; all other groups are fully scored.
\FloatBarrier

\subsection{Pooled Operating Points and Angle-Stratified Results}
Rank-1 and fixed-threshold TAR use all 15,136 attempted queries as the denominator. A query without a score counts as a failure for Rank-1 and TAR, and as a false nonmatch for FNMR. EER and DET curves use successfully scored genuine and impostor pairs. Thus DMD's EER/DET denominator excludes its single unscored query, whereas its reported Rank-1, fixed-threshold TAR, and FNMR include that failure.

For each matcher, one threshold is selected from the pooled impostor scores of all angles to give an overall FAR $\le$ 0.1\% (Table~\ref{tab:supp_projection_operating}) and is applied to every angle group (Tables~\ref{tab:supp_projection_dmd}--\ref{tab:supp_projection_vf}); per-group FARs therefore range from 0.05\% to 0.16\%. Figures~\ref{fig:supp_projection_dmd} and~\ref{fig:supp_projection_vf} show angle-stratified DET curves and FNMR at this threshold. Table~\ref{tab:supp_projection_roc} instead selects a threshold for each group from its own ROC using scored pairs only, which is why DMD's all-angle TAR is 98.27\% there and 98.26\% in Table~\ref{tab:supp_projection_operating}.

\begin{table}[!t]
\blacktable
\revisionfloatcolor
\centering\footnotesize
\setlength{\tabcolsep}{3pt}
\caption{\rev{Stage 3 Overall Results and Pooled Operating Points}}
\label{tab:supp_projection_operating}
\begin{tabular}{lrr}
\toprule
Quantity & DMD & VeriFinger\\
\midrule
Attempted queries &15,136&15,136\\
Valid genuine scores &15,135&15,136\\
Rank-1 (\%) &98.57&98.64\\
EER (\%) &0.839&0.805\\
Fixed-threshold TAR (\%) &98.26&98.50\\
Fixed-threshold FNMR (\%) &1.74&1.50\\
Pooled threshold &0.011284821667&29.00000191\\
Actual overall FAR (\%) &0.10000&0.07893\\
Unscored queries &1&0\\
\bottomrule
\end{tabular}
\end{table}

\begin{table}[!t]
\blacktable
\revisionfloatcolor
\centering\scriptsize
\setlength{\tabcolsep}{2.7pt}
\caption{\rev{Stage 3 DMD Results by Roll Angle (Rates in \%)}}
\label{tab:supp_projection_dmd}
\begin{threeparttable}
\begin{tabular}{lrrrrrr}
\toprule
Roll ($^\circ$)&Queries&R-1&EER&TAR&FNMR&FAR\\
\midrule
$-59$ to $-31$&3,378&97.28&1.362&96.48&3.52&0.0646\\
$-30$ to $-10$&3,339&99.88&0.180&99.70&0.30&0.0709\\
$0$&1,712&99.88&0.116&99.88&0.12&0.0956\\
$10$ to $30$&3,340&99.91&0.119&99.88&0.12&0.1047\\
$31$ to $59$&3,367&96.55&1.634&96.20&3.80&0.1620\\
\bottomrule
\end{tabular}
\begin{tablenotes}[flushleft]
\setlength{\labelsep}{0pt}
\scriptsize\raggedright
\item[] TAR, FNMR, and FAR use the single pooled DMD threshold.
\end{tablenotes}
\end{threeparttable}
\end{table}

\begin{table}[!t]
\blacktable
\revisionfloatcolor
\centering\scriptsize
\setlength{\tabcolsep}{2.7pt}
\caption{\rev{Stage 3 VeriFinger Results by Roll Angle (Rates in \%)}}
\label{tab:supp_projection_vf}
\begin{threeparttable}
\begin{tabular}{lrrrrrr}
\toprule
Roll ($^\circ$)&Queries&R-1&EER&TAR&FNMR&FAR\\
\midrule
$-59$ to $-31$&3,378&97.42&1.498&97.07&2.93&0.0492\\
$-30$ to $-10$&3,339&99.52&0.407&99.49&0.51&0.0894\\
$0$&1,712&99.77&0.192&99.77&0.23&0.1170\\
$10$ to $30$&3,340&99.73&0.206&99.76&0.24&0.0937\\
$31$ to $59$&3,367&97.33&1.190&97.06&2.94&0.0644\\
\bottomrule
\end{tabular}
\begin{tablenotes}[flushleft]
\setlength{\labelsep}{0pt}
\scriptsize\raggedright
\item[] TAR, FNMR, and FAR use the single pooled VeriFinger threshold.
\end{tablenotes}
\end{threeparttable}
\end{table}

\begin{table}[!t]
\blacktable
\revisionfloatcolor
\centering\footnotesize
\setlength{\tabcolsep}{3pt}
\caption{\rev{Stage 3 TAR at Group-Specific ROC Operating Points (\%)}}
\label{tab:supp_projection_roc}
\begin{threeparttable}
\begin{tabular}{lrrrr}
\toprule
&\multicolumn{2}{c}{DMD}&\multicolumn{2}{c}{VeriFinger}\\
\cmidrule(lr){2-3}\cmidrule(lr){4-5}
Roll ($^\circ$)&0.1\%&0.01\%&0.1\%&0.01\%\\
\midrule
All&98.27&96.56&98.50&97.60\\
$-59$ to $-31$&96.83&94.37&97.54&95.86\\
$-30$ to $-10$&99.73&99.28&99.49&99.22\\
$0$&99.88&99.65&99.77&99.71\\
$10$ to $30$&99.88&99.34&99.76&99.61\\
$31$ to $59$&95.49&91.80&97.30&94.98\\
\bottomrule
\end{tabular}
\begin{tablenotes}[flushleft]
\setlength{\labelsep}{0pt}
\scriptsize\raggedright
\item[] Column headings specify empirical FAR upper bounds.
\end{tablenotes}
\end{threeparttable}
\end{table}

\begin{figure*}[!t]
\revisionfloatcolor
\centering
\includegraphics[width=.96\textwidth]{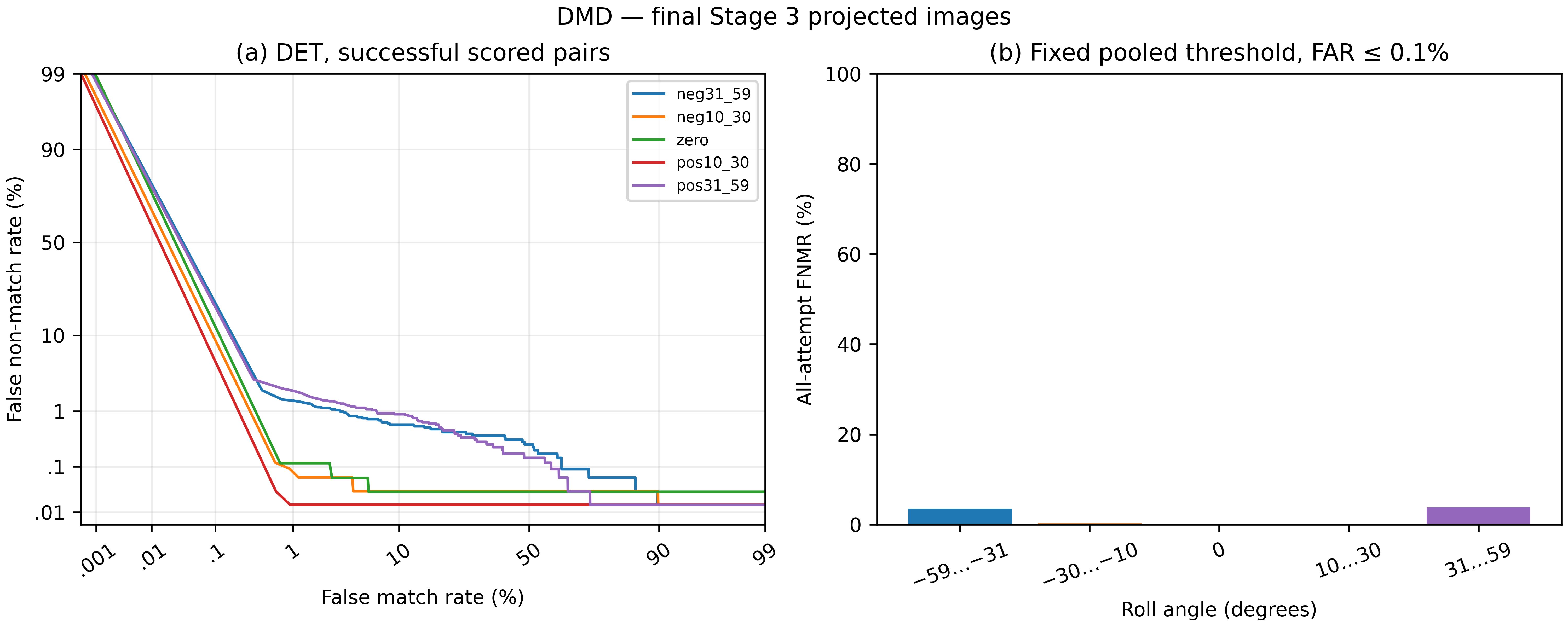}
\caption{\rev{Final Stage~3 DMD evaluation: (a) DET curves by roll group; (b) FNMR at the pooled threshold (overall FAR $\le$ 0.1\%), with all attempted queries in the denominator.}}
\label{fig:supp_projection_dmd}
\end{figure*}

\begin{figure*}[!t]
\revisionfloatcolor
\centering
\includegraphics[width=.96\textwidth]{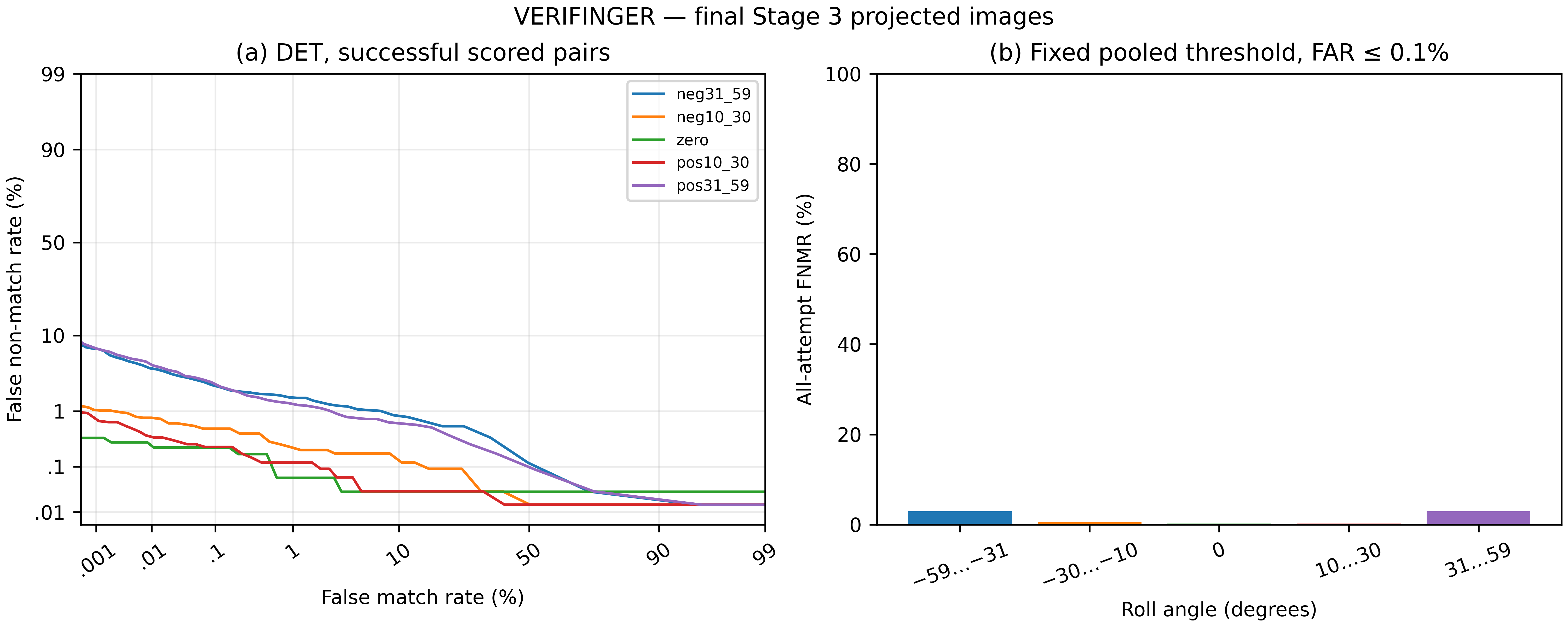}
\caption{\rev{Final Stage~3 VeriFinger evaluation: (a) DET curves by roll group; (b) FNMR at the pooled threshold (overall FAR $\le$ 0.1\%).}}
\label{fig:supp_projection_vf}
\end{figure*}

\subsection{Large-Angle Matching Difficulties}
\label{sec:supp_large_angle}
Both matchers have lower Rank-1 at large absolute roll angles than at $0^\circ$. DMD gives 97.28\% and 96.55\% in the negative and positive extreme groups, versus 99.88\% at $0^\circ$; VeriFinger gives 97.42\% and 97.33\%, versus 99.77\%. Projection thus costs 2.4--3.3 Rank-1 points at large angles relative to the frontal view.

The unscored DMD query, \texttt{05194\_-58.png}, yields only one minutia, below the five required by the DMD/LSAR implementation. Its visible ridge region is markedly narrower than in the frontal view of the same identity, illustrating the visibility loss described in the main paper.
\end{revision}

%% file: sections/supp_training.tex
\begin{revision}
\subsection{Controlled Training Protocol for the Projected-View Ablation}
\label{sec:supp_controlled_training}

The controlled ablation in the main text compares the use of frontal source images and projected source views under the same optimization budget. All three settings use the same 1,712 synthetic identities and the same 1,712 rolled source images. Table~\ref{tab:supp_ablation_sources} counts distinct source files.

\begin{table}[!t]
\blacktable
\revisionfloatcolor
\centering
\footnotesize
\setlength{\tabcolsep}{3pt}
\caption{\rev{Distinct Source Images in the Controlled Ablation}}
\label{tab:supp_ablation_sources}
\begin{tabular}{@{}llrrr@{}}
\toprule
Setting & Contactless source & IDs & Rolled & Contactless \\
\midrule
A & Stage 2 frontal & 1,712 & 1,712 & 1,712 \\
B & Stage 2 frontal & 1,712 & 1,712 & 1,712 \\
C & Stage 3 projected & 1,712 & 1,712 & 15,136 \\
\bottomrule
\end{tabular}
\end{table}

Setting A uses the common affine transformations and noise processing of the training pipeline. Setting B uses the same frontal images and adds principal-component-analysis-based thin-plate-spline (PCA-TPS) deformation on the contactless side. Setting C replaces the frontal contactless source pool with the final Stage 3 projected views and uses the same contactless-side augmentation as B. Rolled-side histogram processing and the other shared augmentations are identical across A/B/C. A is therefore an augmented baseline. The tested difference between B and C is the use of projected source views, including their deformation, visibility, and diversity, relative to the tested two-dimensional augmentation.

Table~\ref{tab:supp_ablation_protocol} records the common training settings. The FDD initialization and the initial states of both training classification heads are identical across A/B/C, with matching SHA-256 hashes for both initialization and identity-presentation order (seed 6131) across all three runs. The descriptor and the two classification heads are trainable, whereas the external pose estimator remains fixed. This protocol is separate from the main text's joint pose-and-descriptor fine-tuning experiment.

\begin{table}[!t]
\blacktable
\revisionfloatcolor
\centering
\footnotesize
\setlength{\tabcolsep}{3pt}
\renewcommand{\arraystretch}{1.12}
\caption{\rev{Shared Settings of the Controlled A/B/C Runs}}
\label{tab:supp_ablation_protocol}
\begin{tabular}{@{}p{0.29\columnwidth}p{0.66\columnwidth}@{}}
\toprule
Item & Setting \\
\midrule
Trainable modules & FDD descriptor and two training classification heads \\
Fixed pose model & \texttt{pose\_guan} \\
Initialization & Identical FDD and head states \\
Sampling & Same identity-presentation order; seed 6131 \\
Optimizer & AdamW; learning rate $3.5\times10^{-4}$; weight decay $4\times10^{-5}$ \\
Numerical precision & FP32 \\
Microbatch & 24 image pairs \\
Gradient accumulation & Two microbatches per update \\
Effective batch & 48 image pairs per update \\
Optimizer updates & 210 per setting \\
Training exposure & 10,080 pair presentations; 20,160 image presentations \\
Checkpoint & Final step 210; no selection by UWA or PolyU test scores \\
\bottomrule
\end{tabular}
\end{table}

At an effective batch of 48 pairs per update, 210 updates correspond to $210\times48=10{,}080$ pair presentations, or 20,160 image presentations, including repeated sampling. The budget is matched in updates and pair presentations, whereas setting C draws on more distinct contactless views (Table~\ref{tab:supp_ablation_sources}). The final checkpoint, \texttt{step\_000210}, is used for every setting without choosing a checkpoint according to performance on either test database.

For all three settings, pose inputs come from the fixed \texttt{pose\_guan} model. FingerNet masks use a threshold of 0.7 without erosion. Minutiae supervision is extracted independently for each image. If minutiae cannot be extracted from a contactless view, its minutiae loss is omitted and the view is kept.

\subsubsection{Common evaluation and interpretation}
The initial FDD and all A/B/C checkpoints are evaluated using one evaluation entry point, identical test lists, fixed pose inputs, the same preprocessing, and masked dense cosine scoring. Table~\ref{tab:supp_test_counts} gives the query, gallery, genuine-pair, and impostor-pair counts. Each metric in the controlled-ablation table (Table~\ref{tab:stage3_ablation}) is computed under this common protocol.

\begin{table}[!t]
\blacktable
\revisionfloatcolor
\centering
\footnotesize
\setlength{\tabcolsep}{3pt}
\caption{\rev{Common Test-Set and Comparison Counts}}
\label{tab:supp_test_counts}
\begin{tabular}{@{}lrrrr@{}}
\toprule
Test set & Queries & Gallery & Genuine & Impostor \\
\midrule
UWA & 2,988 & 1,000 & 2,988 & 2,985,012 \\
PolyU & 2,016 & 2,016 & 12,096 & 4,052,160 \\
\bottomrule
\end{tabular}
\end{table}

Across the three settings, C achieves the best Rank-1 and EER on both databases, while also surpassing the initial FDD in Rank-1 and low-FAR TARs on PolyU.

\subsection{Fine-Tuning with Nominally 500 Identities}
\label{sec:supp_historical_training}

The two runs in Table~\mainref{tab:matched500} use 500 synthetic identities and 498 real-UWA identities with complete image pairs. Table~\ref{tab:supp_historical_budget} lists their source images and training budgets. Both run for six epochs with two accumulated microbatches of 24 pairs, i.e., 60 updates or $60\times48=2{,}880$ pair presentations. The synthetic run contains more contactless images and wider pose coverage than the real run.

\begin{table}[!t]
\blacktable
\revisionfloatcolor
\centering
\footnotesize
\setlength{\tabcolsep}{3pt}
\renewcommand{\arraystretch}{1.1}
\caption{\rev{Source Pools and Budgets of the 500-Identity FDD Runs}}
\label{tab:supp_historical_budget}
\begin{tabular}{@{}p{0.43\columnwidth}rr@{}}
\toprule
Item & Synthetic & Real UWA \\
\midrule
Training identities & 500 & 498 \\
Rolled/contact source images & 500 & 498 \\
Contactless source images & 4,403 & 2,988 \\
Epochs & 6 & 6 \\
Pairs per microbatch & 24 & 24 \\
Accumulation steps & 2 & 2 \\
Optimizer updates & 60 & 60 \\
Update-associated pairs & 2,880 & 2,880 \\
Update-associated images & 5,760 & 5,760 \\
\bottomrule
\end{tabular}
\end{table}

Their initialization has the same SHA-256 hash as the pretrained FDD baseline. The initial and both fine-tuned checkpoints are reevaluated with the test lists, pose inputs, preprocessing, and scoring of Table~\ref{tab:supp_test_counts}; under this protocol the initial PolyU EER is 3.19\% (3.20\% in the original evaluation). Because both runs keep the pose estimator fixed and train the descriptor jointly with two newly initialized classification heads for only 60 updates, early head gradients perturb the pretrained features before convergence, causing both checkpoints to fall below the initial FDD. Even so, IMPOSE data preserves cross-dataset generalization on PolyU much better than real UWA data (Rank-1 $-0.75$ and EER $+0.83$ points vs.\ $-7.15$ and $+4.47$ points), while real UWA data degrades slightly less on UWA.
\end{revision}